\documentclass[10pt,twocolumn,letterpaper]{article}

\usepackage[preprint]{cvpr}      

\usepackage{amsmath,amssymb,amsthm}
\newtheorem{theorem}{Theorem}
\usepackage[utf8]{inputenc} 
\usepackage[T1]{fontenc}    
\usepackage{url}            
\usepackage{booktabs}       
\usepackage{amsfonts}       
\usepackage{nicefrac}       
\usepackage{microtype}      
\usepackage{xcolor}         
\usepackage{graphicx}
\usepackage{subcaption}

\usepackage{titletoc}
\usepackage[titletoc]{appendix}

\newcommand{\concept}{c_t}

\definecolor{cvprblue}{rgb}{0.21,0.49,0.74}
\usepackage[pagebackref,breaklinks,colorlinks,allcolors=cvprblue]{hyperref}

\def\paperID{*****} 
\def\confName{CVPR}
\def\confYear{2026}

\title{Beyond Top Activations: \\ Efficient and Reliable Crowdsourced Evaluation of Automated Interpretability}

\author{
  Tuomas Oikarinen \\
  UC San Diego, CSE \\
  toikarinen@ucsd.edu 
  \and
  Ge Yan \\
  UC San Diego, CSE \\
  geyan@ucsd.edu
  \and
  Akshay Kulkarni \\
  UC San Diego, CSE \\
  a2kulkarni@ucsd.edu
  \and
  Tsui-Wei Weng \\
   UC San Diego, HDSI \\
   lweng@ucsd.edu
}

\begin{document}
\maketitle
\begin{abstract}
Interpreting individual neurons or directions in activation space is an important topic in mechanistic interpretability. Numerous automated interpretability methods have been proposed to generate such explanations, but it remains unclear how reliable these explanations are, and which methods produce the most accurate descriptions. While crowd-sourced evaluations are commonly used, existing pipelines are noisy, costly, and typically assess only the highest-activating inputs, leading to unreliable results. In this paper, we introduce two techniques to enable cost-effective and accurate crowdsourced evaluation of automated interpretability methods beyond top activating inputs. First, we propose Model-Guided  Importance Sampling (MG-IS) to select the most informative inputs to show human raters. In our experiments, we show this reduces the number of inputs needed to reach the same evaluation accuracy by $\sim13\times$. Second, we address label noise in crowd-sourced ratings through Bayesian Rating Aggregation (BRAgg), which allows us to reduce the number of ratings per input required to overcome noise by $\sim3\times$. Together, these techniques reduce the evaluation cost by $\sim40\times$, making large-scale evaluation feasible. Finally, we use our methods to conduct a large scale crowd-sourced study comparing recent automated interpretability methods for vision networks.
\end{abstract}    
\vspace{-6mm}
\section{Introduction}

Despite their transformative capabilities, deep learning models remain fundamentally opaque, which limits their reliability and trustworthiness, especially in high-stakes domains. To address this, the field of mechanistic interpretability aiming to understand the mechanisms inside neural networks has grown rapidly. A key part of mechanistic interpretability is understanding small components of neural networks such as individual neurons or latents in a sparse autoencoder (SAE) through text-based descriptions, which can be achieved via automated neuron descriptions~\cite{bau2020understanding,mu2021compositional,hernandez2022natural,oikarinen2023clip,oikarinen2024linear,bykov2023labeling,la2023towards,shaham2024multimodal,bai2025interpret}.

\begin{figure*}[t!]
    \centering
    \includegraphics[width=.93\textwidth]{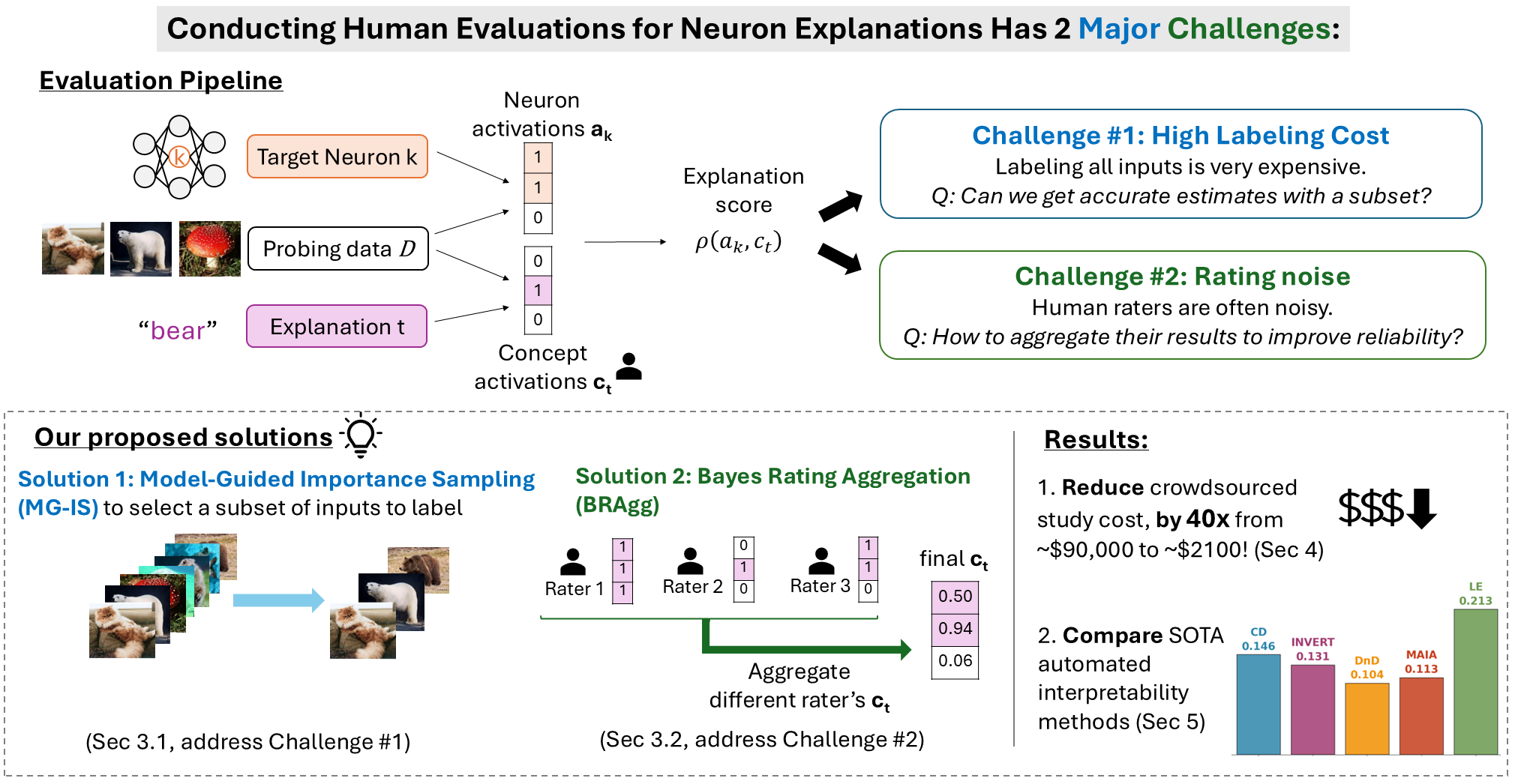}
    \caption{Overview of the explanation evaluation pipeline. We focus on two main challenges: \textcolor{blue}{1: How to reduce high labeling cost?} and \textcolor{Green}{2: How to effectively handle noisy ratings?}. Our proposed solutions are discussed in Section~\ref{sec:method}, validated in Section~\ref{sec:validation}, and applied to compare automated interpretability methods in Section~\ref{sec:large_scale_exp}.}
    \label{fig:overview}
\end{figure*}

To assess the quality of these neuron explanations, researchers often rely on human studies, such as crowd-sourced evaluations via platforms like Amazon Mechanical Turk (AMT). However, despite their importance, the design and analysis of these studies has received relatively little attention. Most existing evaluations examine only whether an explanation matches the top activating inputs of a neuron. As identified by \cite{oikarinen2025evaluating}, this corresponds to measuring only \textit{Recall}, and ignores many important factors such as whether the explanation also describes lower activations of that neuron, or whether all inputs matching the explanation actually activate the neuron (\textit{Precision}). This can lead to misleading conclusions and fails to reliably assess explanation quality. 

Motivated by the above limitations, we aim to conduct a crowdsourced evaluation using a more principled evaluation metric: the \textit{correlation coefficient} between neuron activations and concept (explanation) presence labels. We choose correlation as it was identified as the most reliable metric by \cite{oikarinen2025evaluating} -- it captures both sensitivity and specificity and it doesn't require arbitrary binarization of real-valued neuron activations. However, using correlation as the evaluation metric introduces two challenges that directly drive up total human study cost:
\begin{itemize}
    \item \textbf{Challenge 1: High Labeling Cost}. Computing the correlation requires concept presence labels across the \textit{entire} probing dataset. Labeling all inputs of a dataset with 50,000 images for a single (neuron, explanation) pair with three raters each costs roughly \$600, and evaluating a few thousands of neurons in a single model can reach \$1M. 
    \item \textbf{Challenge 2: Rater Noise}. Human concept judgments are noisy, and even small error rates can significantly affect correlation scores, particularly for rare concepts. Increasing the number of raters per input improves reliability, but further multiplies overall cost.
\end{itemize}
In other words, \textit{reliable, full-distribution human evaluation of neuron explanations is currently too expensive to perform at scale.} To address this, we introduce two novel techniques to enable cost-efficient and reliable crowd-sourced evaluation of automated interpretability.\footnote{Our code will be available at \href{https://github.com/Trustworthy-ML-Lab/Efficient-Interpretability-Eval}{https://github.com/Trustworthy-ML-Lab/Efficient-Interpretability-Eval}} In summary, our contributions are:
\begin{itemize}
    \item \textbf{1. Model-Guided Importance Sampling (MG-IS)} \textit{for Efficient Input Selection}: We develop an importance sampling strategy that selects the most informative inputs to show raters, reducing labeling cost by around 13× while preserving correlation estimation accuracy;
    \item \textbf{2. Bayes Rating Aggregation (BRAgg)} \textit{for Robust Rater Modeling}: We propose a Bayesian rater uncertainty model that reduces the number of ratings required for stable correlation estimates by an additional 3×, which equivalently reduce the labeling cost by 3$\times$;
    \item \textbf{3. Systematic Comparison of Automated Interpretability Methods:} Together, our proposed MG-IS and BRAgg reduce total evaluation cost by around \textbf{40$\times$}, taking the total costs of a systematic comparison down from around $\$90,000$ to $\$2160$. Our work enables a large-scale, systematic comparison of leading neuron explanation methods across multiple vision models. Our findings demonstrate that Linear Explanations \cite{oikarinen2024linear} overall produces the best vision neuron explanations, outperforming for example recent LLM-based explanations \cite{bai2025interpret, shaham2024multimodal}. 
\end{itemize}
\section{Related Work}

\subsection{Automated Interpretability Methods for Deep Vision Models}

\textbf{Methods based on labeled concept datasets:} Early work on automated interpretability sought to match neuron activations to human-interpretable concept annotations. Network Dissection (ND) \cite{netdissect2017} is the first automated method to generate text descriptions for individual neurons in vision models. They introduced the Broden dataset with dense pixel-wise annotations and proposed identifying  concepts with high Intersection over Union (IoU) with binarized neuron activations. However, their approach requires dense concept annotated data which is expensive to collect. Compositional Explanations \cite{mu2021compositional} extends Network Dissection to deal with polysemantic neurons, which may activate on multiple unrelated concepts, by searching for logical compositions of concepts such as "cat OR dog".
Clustered Compositional Explanations(CCE)\cite{la2023towards} extends this even further, addressing the problem that Compositional Explanations only explains the highest activations of a neuron. CCE divides the activation range of a neuron into 5 buckets, and generates a separate compositional explanation for each. \cite{bau2020understanding} propose a version of Network Dissection utilizing a segmentation model instead of labeled data. INVERT\cite{bykov2023labeling} proposes explaining neurons with logical compositions of dataset class(or superclass) labels that maximize AUC, which removes the reliance on Broden but still requires labeled data. 

\noindent \textbf{Methods based on Language Models:} Another popular line of work leverages generative language models to generate natural-language neuron descriptions. The first such method was MILAN \cite{hernandez2022natural}, which trains a small generative neural network to describe the most highly activating inputs of a neuron. Describe-and-Dissect (DnD) \cite{bai2025interpret} is a more recent method that utilizes pre-trained language models instead of training their own to generate detailed neuron descriptions based on highest activating inputs. Multimodal Automated Interpretability Agent (MAIA)\cite{shaham2024multimodal} is another LLM based explanation pipeline where the explaining LLM agent can interact with many tools such as looking at highly activating inputs, generating new images or editing existing ones to generate its description.

\noindent \textbf{CLIP-based methods:} Finally, papers such as CLIP-Dissect(CD) \cite{oikarinen2023clip} and FALCON \cite{kalibhat2023identifying} have proposed methods that don't require labeled concept information by relying on supervision from multimodal models such as CLIP \cite{radford2021learning}. Linear Explanations (LE) \cite{oikarinen2024linear} proposes to explain polysemantic neurons as a linear combination of concepts, such as "$3 \times \text{dog} + 2\times \text{cat}$". To obtain these explanations, the authors utilize either the class labels from the dataset which gives the explanations denoted as LE(label), or pseudo-labels from SigLIP \cite{zhai2023sigmoid} gives gives the explanation denoted as LE(SigLIP).

\subsection{Human Evaluations of Automated Interpretability Methods}
Crowd-sourced human evaluations are commonly used to assess the quality of automatic neuron explanations in previous works~\cite{netdissect2017, oikarinen2023clip, bai2025interpret}. These evaluations typically present raters the top-activating inputs for a neuron along with one or multiple candidate descriptions, and ask them to rate how well the description matches the top-activating inputs. 
However, as pointed out by NeuronEval~\cite{oikarinen2025evaluating}, this evaluation protocol is flawed and fails to measure whether the description matches lower activations of the neuron, or whether all inputs corresponding to description cause the neuron to fire, and in fact corresponds to only measuring \textit{Recall}. As a result, this evaluation approach can favor explanations that are overly broad or non-specific. 

On the other hand, in \cite{borowski2021exemplary} and \cite{zimmermann2023scale}, the authors conducted more objective crowd-sourced evaluations of neuron interpretability by having raters predict whether an input is highly activating or not based on a feature visualization or a set of highly activating images for that neuron. However, these studies did not study text descriptions of neurons, and their methodology, which is roughly equivalent to measuring AUC may also be biased towards overly generic neuron descriptions and does not pass the sanity checks proposed by \cite{oikarinen2025evaluating}.

NeuronEval \cite{oikarinen2025evaluating} formalizes the task of Neuron evaluation under a unified mathematical framework which we utilize in this paper, and identifies flawed and good evaluation metrics through sanity tests. However, they do not conduct actual user studies using an evaluation metric that pass their sanity checks, which is the goal of this paper.

\section{Methods}
\label{sec:method}

In this section, we introduce two techniques to enable cost-efficient and reliable crowdsourced evaluation of neuron explanations. Section \ref{subsec:challenge_1} introduces Model-Guided Importance Sampling (MG-IS), which addresses the labeling cost challenge by selecting the most informative inputs to label. Section \ref{subsec:challenge_2} introduces Bayes Rating Aggregation (BRAgg), which addresses the rater noise challenge by modeling rater errors to robustly aggregate binary judgments. Together, MG-IS and BRAgg substantially reduce the number of labels and ratings required for stable correlation estimates, making crowdsourced evaluation feasible in practice.

\textbf{Problem Formulation.} Given a neural network $f$ and a neuron of interest $k$ (or a direction in activation space), we can obtain a neuron explanation $t$ (described by text) either manually or via automated interpretability methods, that aims to describe the role of that neuron. 
To measure how good this explanation is, we follow the framework introduced by \cite{oikarinen2025evaluating}, and use the Pearson correlation between the neuron activation vector $a_k$ and the explanation concept presence vector $c_t$ (see Fig. \ref{fig:overview} for an overview of the evaluation pipeline). 
A high correlation coefficient means the concept is present on inputs where the neuron activates highly, and not present when neuron activates lowly.

Here, $a_k \in \mathbb{R}^{|\mathcal{D}|}$ is the activation vector of a neuron $k$ over all the probing images $x_i \in \mathcal{D}$ and $\concept \in \mathbb{R}^{|\mathcal{D}|}$ is the concept vector to indicate if concept $t$ appears on the probing images. The $i$-th component $[a_k]_{i}$ and $[\concept]_{i}$ can be formally expressed as:
$[a_k]_{i} = f^{0:l}_k(x_i), \;\; [\concept]_{i} = \mathbb{P}(t|x_i),\;\;  \forall \; i \in \{|\mathcal{D}|\}$ where $f^{0:l}$ denotes the neural network up to $l$-th layer. Our goal is to evaluate Pearson's correlation coefficient $\rho$, which can be expressed as:
\begin{equation}
\label{eq:s_corr}
    \rho(a_k, c_t) = \frac{1}{|\mathcal{D}|} 
             \frac{\sum_{i \in \mathcal{D}} ([a_k]_i - \mu(a_k)) \cdot ([\concept]_i - \mu(\concept))}{\sigma(a_k)\sigma(\concept)}
\end{equation}
where $\mu(a_k), \sigma(a_k), \mu(\concept), \sigma(\concept)$ are the mean and standard deviation of the vectors $a_k$ and $\concept$. 

\subsection{Model-Guided Importance Sampling (MG-IS)}
\label{subsec:challenge_1}

The activation vector $a_k$ can be easily evaluated with a forward pass of the neural network. However, obtaining a high quality concept vector $\concept$ can be expensive and difficult, requiring human raters or computationally expensive models. Our problem then becomes: \textit{How can we best select a subset of inputs $S \subseteq \{|D|\}$ to label, such that we can accurately estimate $\rho(a_k, c_t)$ using just the subset?}

The simplest approach is to select input for annotation using uniform (Monte Carlo) sampling, where each input has an equal probability of being selected. However, concepts of interest could be rare in the probing dataset, meaning that a uniform sample is unlikely to contain sufficient amount (or any) of positive examples of the concept, making accurate estimation of correlation impossible.



\subsubsection{Method: MG-IS}
To address this issue, we propose estimating $\rho$ using importance sampling, where we draw inputs $x_i$ from a proposal distribution $\mathcal{Q}$ that places higher probability on informative inputs such as the ones that are likely to contain the concept, rather than from the original uniform distribution $\mathcal{P}$. Importantly, importance sampling preserves the expectation we seek to estimate, i.e. $ \mathbb{E}_{x \sim \mathcal{P}}[h(x)] =  \mathbb{E}_{x \sim \mathcal{Q}}[\frac{p(x)}{q(x)}h(x)]$ for any distribution $Q$, given $q(x) > 0$ for all $x$ where $p(x)h(x) \neq 0$. Here $p(x)$ and $q(x)$ are the probability mass functions (pmf) of $\mathcal{P}$ and $\mathcal{Q}$ respectively. 

According to \cite{montecarlobook} (Sec 3.3.2, Theorem 3.12, reproduced in Appendix \ref{sec:IS_thm}), when estimating the expected value of a function $h$, i.e. $\mathbb{E}_{x \sim \mathcal{P}} [h(x)]$, the optimal proposal distribution $\mathcal{Q}^*$ that minimizes the variance of the estimator should have the pmf $q^*(x) \propto |h(x)|p(x)$. Intuitively, this distribution samples more frequently from inputs that make a larger contribution to the value of the expectation, reducing the number of samples needed for a stable estimate.

To apply importance sampling to our correlation estimation, we first rewrite \eqref{eq:s_corr} in the form of an expectation over inputs: 
\begin{equation}
\label{eq:s_corr_p}
    \rho = \mathbb{E}_{x_i \sim \mathcal{P}}
             \frac{([a_k]_i - \mu(a_k)) \cdot ([\concept]_i - \mu(\concept))}{\sigma(a_k)\sigma(\concept)},
\end{equation}
Based on the Theorem~\ref{thm:IS} and Eq~\eqref{eq:s_corr_p}, we can  estimate the $\rho$ by sampling $x_i$ from $\mathcal{Q}^*$ with pmf $q(x)$ as:
\begin{equation}
\label{eq:opt_q_sample}
    q^*(x_i) \propto \left|\frac{([a_k]_i - \mu(a_k)) \cdot ([\concept]_i - \mu(\concept))}{\sigma(a_k)\sigma(\concept)}\right| \cdot \frac{1}{|\mathcal{D}|}
\end{equation}
\begin{equation}
    \implies q^*(x_i) \propto |\bar{a}_{ki} \cdot \bar{c}_{ti}|,
\end{equation}
where $\bar{a}_{ki}$ and $\bar{c}_{ti}$ are the normalized quantities $\frac{[a_k]_i - \mu(a_k)}{\sigma(a_k)}$ and $\frac{[\concept]_i - \mu(\concept)}{\sigma(\concept)}$ respectively. So to minimize the variance of estimator $\rho$, we should sample $x_i$ more from the probing images that have high absolute product values $|\bar{a}_{ki} \cdot \bar{c}_{ti}|$. 

Since we do not know $\concept$ before running the test (i.e. through human study or computationally expensive model), we instead use a cheaper (and less accurate) method to approximate $\concept$ for the purposes of sampling. In particular, we use the SigLIP \cite{zhai2023sigmoid} model to predict $\concept$, which we denote as $\concept^{siglip}$. This gives us the $q^{siglip}$ sampling distribution:
\begin{equation}
    q^{siglip}(x_i) = \frac{|[\bar{a}_{k}]_i \cdot [\bar{c}_{t}^{siglip}]_i|}{\sum_{x_i \in \mathcal{D}} |[\bar{a}_{k}]_i \cdot [\bar{c}_{t}^{siglip}]_i|}
\end{equation}

To account for the error in siglip estimates, and to ensure that importance sampling is unbiased ($q(x) > 0$ when $p(x)h(x) \neq 0$), our final sampling distribution $q^{\text{MG-IS}}(x)$ is a mixture of $q^{siglip}(x)$ and the original distribution $p(x) = \frac{1}{|\mathcal{D}|} \ \forall \ x \in \mathcal{D}$:
\begin{equation}
    q^{\text{MG-IS}}(x) = (1-\gamma) q^{siglip}(x) + \gamma p(x)
    \label{eq:mg_is}
\end{equation}

The hyperparameter $\gamma$ was chosen experimentally, with best performance using $\gamma=0.2$. See Appendix \ref{app:gamma_choice} for details. This is equivalent to selecting 80\% of the samples from $q^{siglip}$ and 20\% of them from the uniform distribution. We name this approach model-guided importance sampling (MG-IS) because the sampling is guided by the SigLIP models.

\textbf{Sampling Correction.}
To obtain an unbiased estimate of the correlation using importance sampling, we must estimate not only the final coefficient, but also the mean and variance of the concept vector $c_t$. To evaluate them we apply the importance sampling correction $\frac{p(x)}{q(x)}$ sequentially. We first estimate the mean $\mu_S(\concept)$, and then the standard deviation $ \sigma_S(\concept)$ from the samples $x_i \sim \mathcal{Q}$:
\begin{equation}
    \mu_S(\concept) = \frac{1}{|S|} \sum_{i \in S} \frac{p(x_i)}{q(x_i)} [\concept]_i
\end{equation}
\begin{equation}
    \sigma_S(\concept) = \sqrt{\frac{1}{|S|-1} \sum_{i \in S} \frac{p(x_i)}{q(x_i)} ([\concept]_i-\mu_S(\concept))^2}.
\end{equation}
Next, we normalize the inputs: $[\bar{\concept}]_S = \frac{[\concept]_S-\mu_S(\concept)}{\sigma_S(\concept)}$. For $a_k$, since we have access to the entire neuron activation vector, we can directly compute the true mean and standard deviation (i.e. $\mu(a_k)$, $\sigma(a_k)$) instead of needing to estimate them from the sample. Thus, we have $[\bar{a}_k]_{S} = \frac{[a_k]_{S}-\mu(a_k)}{\sigma(a_k)}$. Finally, we obtain the estimated correlation score of $\rho_{S}$ as:
\begin{equation}
    \rho_S = \frac{1}{|S|} \sum_{i \in S} \frac{p(x_i)}{q(x_i)} [\bar{a}_k]_{i} \cdot [\bar{\concept}]_i,
\end{equation}
where we use the subscript $S$ to denote that the correlation score is estimated on the subset $S$.

\subsubsection{Simulation study: Sampling Effectiveness}
\label{subsec:sampling_simulation}

To study the accuracy of our estimated correlation $\rho_S$, we simulate a human study using existing labeled data.  We then simulate limited data by only using a subset of these labels, and compare different sampling methods based how close their estimated correlation coefficient $\rho_S$ is to the true correlation coefficient $\rho_{gt}$ in Eq. \eqref{eq:rho_gt}. 

In particular, we use the ImageNet\cite{deng2009imagenet} dataset and its class and superclass labels, which we denote as the ground truth concept labels $c^*_t$. 
We then first generate descriptions $t_k$ for all 2048 neurons in layer4 of ResNet-50\cite{he2016deep} trained on ImageNet, by selecting the concept $t_k$ from ImageNet class and superclass names $\mathcal{C}$ that maximizes correlation with neuron $k$'s activations. 
\begin{equation}
\label{eq:neuron_explanation_from_gndtruth}
    t_k = \text{argmax}_{t \in \mathcal{C}} \,\, \rho(a_k, c_t^*) 
\end{equation}
\begin{equation}
    \label{eq:rho_gt}
    \rho_{gt} = \rho(a_k, c_{t_k}^*)
\end{equation}
To compare different sampling strategies, we sample subsets of observations using their proposed distributions $Q$, and measure the Relative Correlation Error (RCE, Eq. \ref{eq:rce}) compared to correlation calculated from full distribution:
\begin{equation}
\label{eq:rce}
    RCE(Q, |S|) = \frac{\sum_{k \in K}\mathbb{E}_{S \sim Q}[|\rho_S(a_k ,c_{t_k}) - \rho_{gt}(a_k, c_{t_k}^{*})|]}{\sum_{k \in K} |\rho_{gt}(a_k, c_{t_k}^{*})|}
\end{equation}
For Uniform Sampling, $Q = P$. In practice we evaluate the expectation by averaging over 10 samples from $Q$.

\begin{figure}
    \centering
    \includegraphics[width=0.4\textwidth]{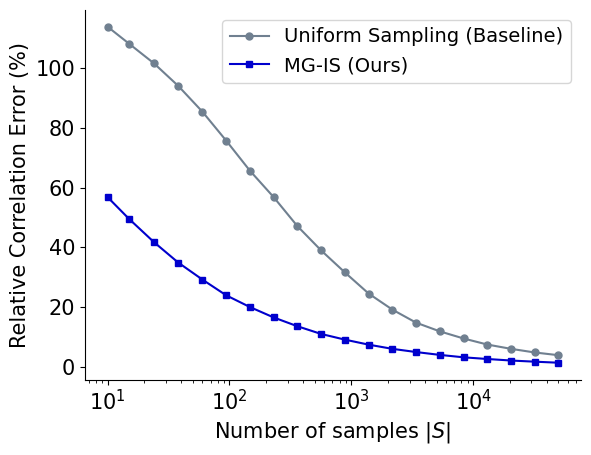}
    \caption{Comparing different sampling strategies to estimate $\rho_{\mathcal{S}}$. Our Model-Guided Importance Sampling (\textcolor{blue}{MG-IS}) using SigLIP estimates significantly outperform baseline regardless of the size of sampled subset.}
    \label{fig:sampling_simulation}
\end{figure}

\textbf{Results.}  Figure \ref{fig:sampling_simulation} shows the Relative Correlation Error of different strategies as a function of number of samples $|S|$. 
We can see that overall our \textcolor{blue}{MG-IS}, as defined in Eq.~\eqref{eq:mg_is}, performs much better -- which has low error and lower cost. Compared to the \textcolor{darkgray}{Uniform Sampling} baseline, we can match the correlation estimation accuracy with around $15 \times$ less samples. Alternatively, we can reach $\sim65\%$ lower correlation estimation error with the same budget.

In Appendix \ref{app:act_guide_is} we also experiment with a version of importance sampling without model guidance, based on $a_k$ only, and show it can still be useful (better than uniform sampling) for the cases where no cheap model is available, but overall it performs worse than the MG-IS.

\subsection{Bayes Rating Aggregation (BRAgg)}
\label{subsec:challenge_2}

Another challenge in human evaluation is that the concept labels $\concept$ obtained are noisy due to variability in rater attention, interpretation, and experience -- this is particularly a problem when using crowd-sourced platforms such as Amazon Mechanical Turk. Such noise can significantly affect correlation estimation, particularly when the concept is rare as we might end up with more false positives than true positives. 

To mitigate this, for each input $x_i$ and concept $t$, we collect $m$ independent binary ratings $r_{ti}^j \in \{0,1\}, j \in \{m\}$ from different raters. Let $R_{ti} = \{r^1_{ti}, ..., r^m_{ti}\}$ denote the set of the ratings for a particular (input, concept) pair $(x_i, t)$. The goal of rating aggregation is to combine these $m$ judgments for each input $x_i$ into a single estimated concept presence value $[c_t]_i$. Below, we describe three different methods for Rating Aggregation to get $[c_t]_i$.


\noindent \textbf{Method 1 - Average:} 
$[\concept]_i = \frac{\sum_{j=1}^m r_{ti}^j}{m}$

$\Rightarrow$ Ratings are simply averaged to form $c_t$.

\noindent \textbf{Method 2 - Majority Vote:}  
    $$[\concept]_i = \begin{cases}1, \quad \text{if} \quad \frac{\sum_{j=1}^m r_{ti}^j}{m} > 0.5; \\
    0, \quad\text{if} \quad \frac{\sum_{j=1}^m r_{ti}^j}{m} \leq  0.5. \\
    \end{cases}
    $$
    
$\Rightarrow$ The most common rating becomes the label $[\concept]_i$.\newline

While these methods are simple and commonly used, they may not be the optimal ways to model our confidence that a concept is present in a particular input. To better capture this uncertainty, we propose Bayes Rating Aggregation (BRAgg) to estimate $[\concept]_i$ as the posterior probability that the concept is present: $[\concept]_i = \mathbb{P}([\concept^*]_i = 1 | R_{ti})$. Here $c^{*}_{ti}$ denotes the "ideal" or true concept value without any labeling noise.   
\begin{figure}
    \centering
    \includegraphics[width=0.48\textwidth]{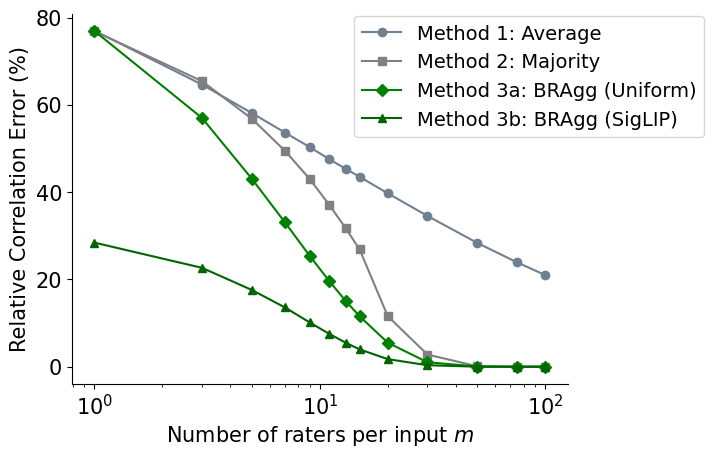}
    \caption{Comparison of different rating aggregation strategies on simulated human study with error rate $\eta=23\%$, tested on the full dataset.}
    \label{fig:rating_aggregation_simulation}
\end{figure}

\noindent \textbf{Method 3 -  Bayes Rating Aggregation (BRAgg):}

Let $C_{ti} := ([\concept^*]_i = 1)$ and $\lnot{C_{ti}} = ([\concept^*]_i = 0])$, then:
\begin{equation}
\label{eq:bayes}
    [\concept]_i = \frac{\mathbb{P}(R_{ti} | C_{ti})\cdot \mathbb{P}(C_{ti})}{\mathbb{P}(R_{ti} | C_{ti}) \cdot \mathbb{P}(C_{ti}) + \mathbb{P}(R_{ti} | \lnot C_{ti})\mathbb{P}(\lnot C_{ti})}
\end{equation}
where we first use Bayes rule to expand the posterior $[\concept]_i = \mathbb{P}(c_{ti}^* = 1 | R_{ti})$ to get Eq~\eqref{eq:bayes}, and then calculate the Likelihood and Prior terms in Eq~\eqref{eq:bayes} as follows:

\noindent \textbf{(I) Likelihood: $\mathbb{P}(R_{ti} | C_{ti})$.} 
We assume each rater makes errors uniformly at random with error rate $\eta$, i.e. for any input $\mathbb{P}(r^j_{ti} = c_{ti}^{*}) = 1-\eta$, where $\eta$ is a parameter we can estimate experimentally. Let $\alpha_{ti} = \sum_{j=1}^m r_{ti}^j$, we obtain the likelihood in below equations:
\begin{equation}
    \mathbb{P}(R_{ti} | C_{ti}) = (1-\eta)^{\alpha_{ti}}(\eta)^{(m-\alpha_{ti})}
\end{equation}
\begin{equation}
    \mathbb{P}(R_{ti} | \lnot C_{ti}) = (\eta)^{\alpha_{ti}}(1-\eta)^{(m-\alpha_{ti})}
\end{equation}
Note we also tested more advanced modeling where different (input, concept) pairs can have different error rates to account for some inputs being more ambiguous/harder to label, but we found that simple uniform error modeling performed slightly better overall. See Appendix \ref{app:error_model_comparison} for full results.

\noindent \textbf{(II) Prior: $\mathbb{P}(C_{ti})$.} There are multiple ways of choosing priors, reflecting the confidence of whether a concept exists in an input $x_i$. In our analysis, we consider two different priors:
\begin{itemize}
    \item \textbf{(a) Uniform Prior:} for all concepts $t$ and all inputs $x_i$, set
    $\mathbb{P}(C_{ti}) = \beta$
    
    \item \textbf{(b) SigLIP Prior:} we leverage knowledge from a cheap evaluator, namely SigLIP to initialize the prior and set
    $\mathbb{P}(C_{ti}) = [c^{siglip}_{t}]_i$
\end{itemize}
Here $\beta$ is a hyperparameter, which is set as $0.05$. We evaluate the effects of different choices for $\beta$ in Appendix \ref{app:beta_ablation}. For the SigLIP prior, we clip the prior to be between $0.001$ and $0.999$ to avoid extreme values which may dominate the final result. Our method with SigLIP prior can be seen as a hybrid evaluation that combines both human and model knowledge.

\subsubsection{Simulation: Effect of Rating Aggregation Method}

\paragraph{Setup:} We follow the ImageNet based setup described in Section \ref{subsec:sampling_simulation} for defining the ground truth correlation $\rho_{gt}$ (Eq. \ref{eq:rho_gt}). However, instead of using a subset of inputs, in this test we evaluate how accurately we can estimate the correlation with access to $m$ noisy ratings for each input. For our testing, we used an error rate $\eta=23\%$, i.e. randomly flipped $23\%$ of the labels, which aligns with our experimental observations. We report the Relative Correlation Error (RCE) between estimated correlations and $\rho_{gt}$ in Figure \ref{fig:rating_aggregation_simulation}.

As we can see from Figure \ref{fig:rating_aggregation_simulation}, Method 3b \textcolor{Green}{BRAgg(SigLIP)} resulted in clearly the lowest RCE, allowing us to reach relatively accurate estimates with just a few raters, while other methods require $>10$ raters per input before reaching low error rates. Overall BRAgg (Uniform) performs the second best, while Average rating aggregation fails to reach low errors even with 100 raters per input. Depending on the desired error rate, our \textcolor{Green}{BRAgg(SigLIP)} can achieve it with $2$-$10 \times$ less raters than the best baseline (majority).

\section{Experimental Validation and Human Study Design: Combining MG-IS and BRAgg}
\label{sec:validation}

\begin{figure*}
    \centering
    \begin{subfigure}{0.45\textwidth}
        \centering
        \includegraphics[width=\textwidth]{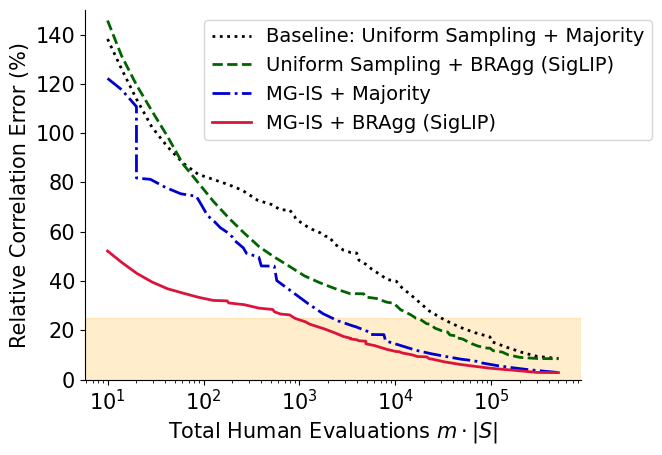}
        \caption{Setting 1: Simulation with 23\% label noise.}
        \label{fig:simulation_combined}
    \end{subfigure}
    \hspace{1cm}
    \begin{subfigure}{0.45\textwidth}
        \centering
        \includegraphics[width=\textwidth]{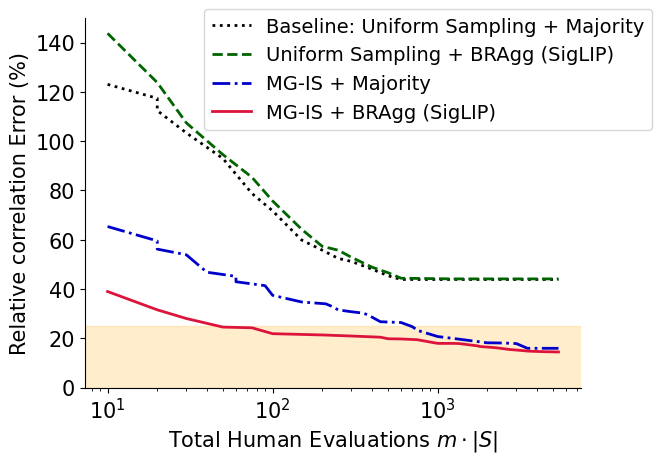}
        \caption{Setting 2: Mechanical Turk method validation.}
        \label{fig:mturk_validation}
    \end{subfigure}
    \caption{Comparing the effect of our sampling technique MG-IS, rating aggregation method BRAgg independently and both together. The shaded region represents RCE$\leq 25\%$.}
    \label{fig:validation_both}
\end{figure*}

To further understand the performance of our proposed techniques, and to determine optimal parameters for our large scale human study in Section \ref{sec:large_scale_exp}, we tested our methods in two realistic settings with both label noise and limited samples. 

\noindent \textbf{Setting 1 - Simulation:} Our first setting is based on the simulation setups described in Section \ref{sec:method}, but this time with \textit{both} reduced samples and 23\% label noise. The results of this setting are shown in Figure \ref{fig:simulation_combined}.

\noindent \textbf{Setting 2 - Crowd-sourced Validation Study:}
In this setting, we conducted a \textit{real} crowd-sourced evaluation on AMT to annotate the concept labels $c_t$ for samples $x_i$ in the probing dataset, on inputs where we also have access to the ground-truth labels $c_t^{*}$ from the dataset for comparison. Specifically, we performed a crowdsourced evaluation of 10 neurons in Resnet-50 layer4 with explanations selected from ImageNet classes and superclasses following Eq.~\eqref{eq:neuron_explanation_from_gndtruth}. We then estimated the correlation coefficient $\rho_S$ using human ratings $R$ on the rated subset $S$ and measured how close our estimate is to the ground truth $\rho_{gt}$. This setup allows us to evaluate the error rate on Mechanical Turk and the effectiveness of our techniques on real crowdsourced data.

We randomly selected 10 neurons (excluding neurons with very vague concepts like \textit{artifact}) and labeled 600 inputs per neuron with 9 AMT raters labeling each input. To estimate the evaluation accuracy with smaller number of raters/inputs we randomly sampled subsets of the ratings for our plots in Figure~\ref{fig:mturk_validation} (Note: each point is the average of 1000 random samples of subsets of the same size). We also used this setting estimate error rate of MTurk raters as $\eta = \mathbb{P}(r_{ti}^j \neq [c_t^*]_i)$, giving us an estimate of 23\%, which we used for our simulations. We note this might slightly overestimate the noisiness of raters as some ImageNet labels are in reality ambiguous or even incorrect but it serves as a useful approximation.

\textbf{Results:} Figure \ref{fig:validation_both} shows the results of different sampling and rating aggregation techniques across our two settings. The x-axis of Figure \ref{fig:validation_both} shows the total number of ratings i.e. number of samples $|S|$ times number of ratings per input $m$. Since the optimal number of ratings per input depends on the total budget (for smaller budgets it is better to use fewer raters per sample), we plot the Pareto-front i.e. the optimal number of raters for each budget in Figure \ref{fig:validation_both}. 

From the results we can see that by far the best results are achieved when using both MG-IS and BRAgg together (\textcolor{red}{red line}), while they also help individually with importance sampling being more important overall. In our Simulation setting (Fig. \ref{fig:simulation_combined}, with a budget of 550 Evaluations per neuron, we can reach RCE of 27.5\%  with \textcolor{red}{MG-IS + BRAgg}. To reach the same error rate requires 1760 ($3\times$) evaluations with \textcolor{blue}{MG-IS only}, 14607 ($25\times$) samples with \textcolor{Green}{BRAgg only} and 22560 ($40 \times$) samples using the baseline uniform sampling with majority vote, showing our techniques lead to around $40 \times$ cost reduction overall. 

On the MTurk evaluation (Fig. \ref{fig:mturk_validation}), overall error rates are lower, but the trends between methods are similar, with MG-IS + BRAgg reaching 19.8\% RCE within the budget of 550 evaluations per neuron, while MG-IS only requires 2000 evals to reach the same error ($4\times$). Uniform sampling methods never reach sufficiently low errors within the number of samples we evaluated, but extrapolating from current results we estimate it would require 10k-30k evaluations to reach 20\% RCE.

Based on our findings and our budget, we choose to use 3 raters per input and to evaluate 180 inputs per neuron for 540 evaluations per neuron in the large scale study in Section~\ref{sec:large_scale_exp}. As we can evaluate 15 inputs in one task for the cost of \$0.06, this evaluation will cost us $\frac{\$0.06}{15} \cdot 180 \cdot 3 = \$2.16$ per (neuron, explanation) pair. See Appendix \ref{sec:crowdsourced_study_design} for detailed results on the tradeoff between more inputs vs. more ratings per input, as well as how to use these experiments to decide the optimal parameters for our large-scale study. 

\textbf{Accuracy of SigLIP only evaluation.} Since our techniques are assisted by the SigLIP model, it is natural to ask how does it compare against fully automated evaluation? 
Our techniques can be seen as a hybrid method combining the strengths of human ratings and efficiency of model-based evaluation. Overall, our results indicate SigLIP-based evaluation is quite reliable by itself. When measuring the RCE of SigLIP only (using the full dataset), it reaches RCE of 32.10\% on all neurons (vs. our estimated human study RCE of 27.90\%), and on the subset of 10 neurons used in our MTurk validation SigLIP only RCE is 22.47\% compared to 19.8\% of our human study. So overall SigLIP only evaluation has around $15\%$ higher error than our human study. This is still more accurate than a low budget human-study, and as such can be a good practical choice for future studies. However, it is important to remember that human evaluation is the gold standard for interpretability, and we should not blindly rely on automated evaluation without validating them with experiments such as ours. See Appendix \ref{app:siglip_accuracy} for a more detailed analysis of SigLIP evaluation accuracy.
\section{Large-Scale Crowdsourced Study}
\label{sec:large_scale_exp}

\begin{table*}[h!]
\centering
\scalebox{0.86}{
\begin{tabular}{@{}lcccccccc@{}}
\toprule
\multicolumn{1}{c}{Simple Exp. ($l$=1)} & 
\begin{tabular}[c]{@{}c@{}}ND\\ \cite{bau2020understanding}\end{tabular}
 & \begin{tabular}[c]{@{}c@{}}MILAN\\ \cite{hernandez2022natural}\end{tabular}
  & \begin{tabular}[c]{@{}c@{}}CD\\ \cite{oikarinen2023clip}\end{tabular}
 & \begin{tabular}[c]{@{}c@{}}INVERT\\ l=1~\cite{bykov2023labeling}\end{tabular} & 
 \begin{tabular}[c]{@{}c@{}}DnD\\ \cite{bai2025interpret}\end{tabular}
  & \begin{tabular}[c]{@{}c@{}}MAIA\\ \cite{shaham2024multimodal}\end{tabular}
   & \begin{tabular}[c]{@{}c@{}}LE(label) \\ l=1~\cite{oikarinen2024linear}\end{tabular} & \begin{tabular}[c]{@{}c@{}}LE(SigLIP)\\ l=1~\cite{oikarinen2024linear}\end{tabular} \\ \midrule
\begin{tabular}[c]{@{}l@{}}RN-50 \\ (Layer4)\end{tabular} & \begin{tabular}[c]{@{}c@{}}0.1242 \\ $\pm$ 0.002\end{tabular} & \begin{tabular}[c]{@{}c@{}}0.0920 \\ $\pm$ 0.002\end{tabular} & \begin{tabular}[c]{@{}c@{}} \underline{0.1904} \\ \underline{$\pm$ 0.002} \end{tabular} & \begin{tabular}[c]{@{}c@{}}0.1867 \\ $\pm$ 0.002\end{tabular} & \begin{tabular}[c]{@{}c@{}}0.1534 \\ $\pm$ 0.002\end{tabular} & \begin{tabular}[c]{@{}c@{}}0.1396 \\ $\pm$ 0.009\end{tabular} & \begin{tabular}[c]{@{}c@{}}0.1793 \\ $\pm$ 0.002\end{tabular} & \textbf{\begin{tabular}[c]{@{}c@{}}0.2413 \\ $\pm$ 0.002\end{tabular}} \\ \midrule
\begin{tabular}[c]{@{}l@{}}ViT-B-16\\ (Layer11 MLP)\end{tabular} & \begin{tabular}[c]{@{}c@{}}0.0335 \\ $\pm$ 0.002\end{tabular} & \begin{tabular}[c]{@{}c@{}}0.0194 \\ $\pm$ 0.002\end{tabular} & \begin{tabular}[c]{@{}c@{}}0.1849 \\ $\pm$ 0.004\end{tabular} & \begin{tabular}[c]{@{}c@{}}0.1343 \\ $\pm$ 0.004\end{tabular} & \begin{tabular}[c]{@{}c@{}}0.1049 \\ $\pm$ 0.003\end{tabular} & \begin{tabular}[c]{@{}c@{}}0.1497 \\ $\pm$ 0.021\end{tabular} & \begin{tabular}[c]{@{}c@{}}\underline{0.2704} \\ \underline{$\pm$ 0.004}\end{tabular} & \textbf{\begin{tabular}[c]{@{}c@{}}0.2968 \\ $\pm$ 0.004\end{tabular}} \\ \bottomrule
\end{tabular}}
\caption{SigLIP based simulation with correlation scoring comparing different simple explanation methods ($l=1$). We can see Linear Explanation (LE) performs the best even when restricted to single concept explanations.}
\label{tab:siglip_sim_len1}
\end{table*}

\begin{figure*}
    \centering
    \begin{subfigure}{0.42\textwidth}
        \centering
        \includegraphics[width=\textwidth]{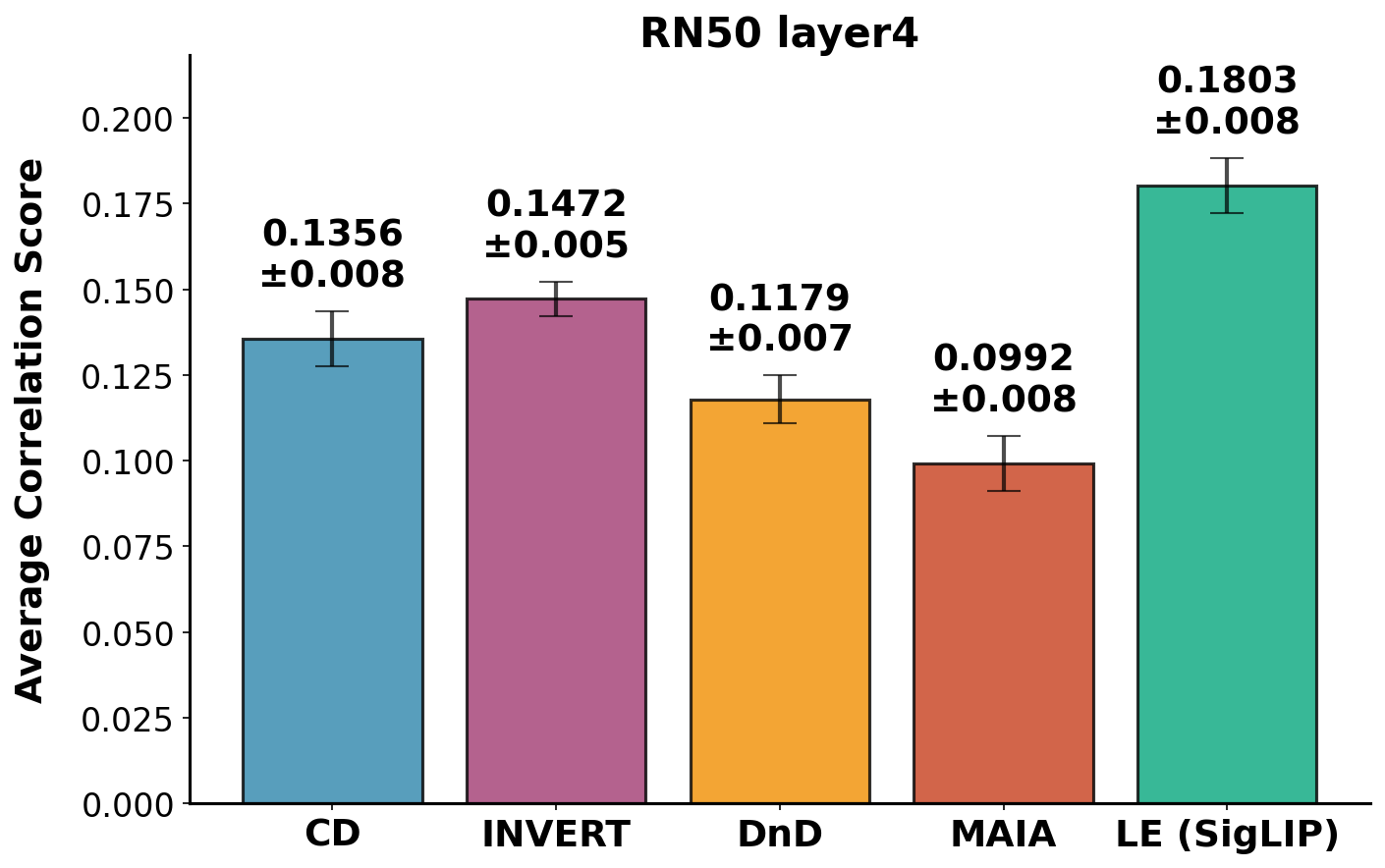}
    \end{subfigure}
    \hspace{1cm}
    \begin{subfigure}{0.42\textwidth}
        \centering
        \includegraphics[width=\textwidth]{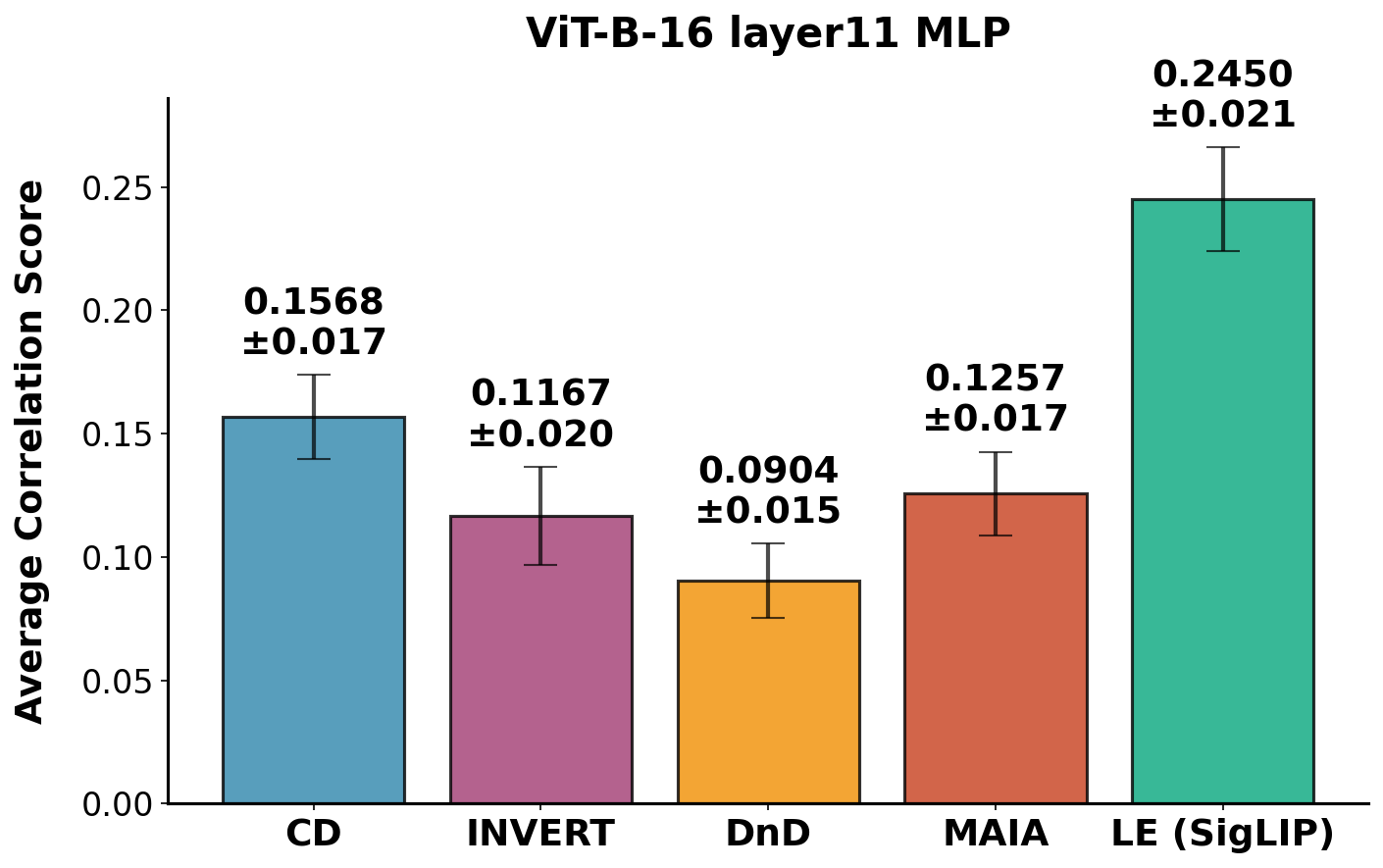}
    \end{subfigure}
    \caption{Results of our MTurk study. Using BRAgg(SigLIP) for rating aggregation. All methods are restricted to length 1 explanations. Overall explanations of LE(SigLIP) have the highest correlation coefficient.}
    \label{fig:large_scale_results}
\end{figure*}

\noindent \textbf{Setup.} Our study focuses on neurons in two layers of two different models: Layer4 of ResNet-50 trained on ImageNet, and MLP activations in Layer11 of ViT-B-16 \cite{dosovitskiy2020image} trained on ImageNet. We use the full ImageNet Validation set as probing dataset, see Appendix \ref{sec:expt_details} for additional details.

\subsection{Method Selection via Automated Evaluation}
\label{subsec:automated_eval}

Many different methods have been introduced to explain neurons in vision models. Due to limited budget, we are unable to compare all these methods on our crowdsourced study, so we first perform an automated evaluation to find the best methods to include in the crowdsourced study. 

\textbf{Explanation Complexity:} Some methods explain a single neuron as a combination of concepts. For example \cite{mu2021compositional, bykov2023labeling} use a logical composition of explanations, such as "dog OR cat". For fairness and simplicity, in our main experiments we focus on explanations with only one concept $l=1$, but we also conduct an automated evaluation of complex experiments in Appendix \ref{app:complex_explanations}.

For automated evaluation we use the SigLIP based simulation with correlation scoring introduced by \cite{oikarinen2024linear}, see Appendix \ref{app:automated_eval_details} for details. The results of this automated evaluation are shown in Table \ref{tab:siglip_sim_len1}.

Overall we can see Linear Explanations \cite{oikarinen2024linear}(even when restricted to a single concept) produces the highest correlation scores.  Other well performing methods include INVERT \cite{bykov2023labeling} and CLIP-Dissect \cite{oikarinen2023clip}, followed by recent language model based explanations MAIA \cite{shaham2024multimodal} and DnD\cite{bai2025interpret}. We observed that Broden based methods \cite{netdissect2017, mu2021compositional, la2023towards} overall performed relatively poorly, likely because their concept sets do not include the relevant higher level concepts such as animal species needed to describe later neurons of ImageNet models.

\subsection{Crowdsourced Evaluation}

\noindent \textbf{Details:} Based on automated evaluation results, we chose to conduct our crowdsourced study on the 5 best performing simple explanations: LE (SigLIP)\cite{oikarinen2024linear}, CLIP-Dissect\cite{oikarinen2023clip}, INVERT\cite{bykov2023labeling}, MAIA\cite{shaham2024multimodal} and DnD\cite{bai2025interpret}. We evaluated 100 randomly selected neurons for each of the two models, and 180 inputs per neuron with 3 raters each. This gives us a total of 1000 (neuron, explanation) pairs to evaluate for a total cost of \$2160. Our study was deemed Exempt from IRB approval by the IRB review board at our institution.

\noindent \textbf{User Interface.} The main objective of our crowdsourced study is to have users annotate which inputs have concept $t$ present. We do this by presenting raters with explanation $t$ and 15 inputs per task, and asking them to select inputs where concept is present. The full study interface and additional details are available in Appendix \ref{app:detailed_mturk_results}.

\subsubsection{Results}

The overall results from the crowdsourced evaluation are shown in Figure \ref{fig:large_scale_results}. See Appendix \ref{app:detailed_mturk_results} for more detailed results, including example explanations and their scores. We can see Linear Explanations performs the best overall despite being limited to using a predefined concept set. We think this is likely because it is the only explanation method tested that is optimized to explain the entire range of activations instead of focusing only on highest activations. The second best performing method is CLIP-Dissect \cite{oikarinen2023clip} or INVERT \cite{bykov2023labeling} depending on the model evaluated. It is somewhat surprising that recent methods based on large language models \cite{bai2025interpret, shaham2024multimodal} did not outperform simpler baselines despite being able to provide very accurate and complex description in some cases. We think this is caused by a few reasons: 
\begin{enumerate}[leftmargin=2em]
    \item Focus on highly activating inputs only, leading to overly specific explanations that do not describe lower activations.
    \item Inconsistency. While all 5 methods mostly produce relevant descriptions, more complex methods such as those relying on LLMs often have more variance in their description quality, leading to poor explanations for some neurons.
\end{enumerate}

Overall the correlation scores were quite low, with best methods reaching $\sim0.2$ correlation, highlighting the need for further improvement in explanation methods and/or creating more interpretable models.
\section{Conclusion}
 In this paper, we introduced two new techniques for conduction principled crowdsourced evaluation of neuron explanations. First, we proposed Model-Guided Importance Sampling (MG-IS) to determine which inputs to show raters, reducing our labeling cost by $\sim15\times$. Second, we introduced a new Bayes rule based method (BRAgg) to aggregate multiple ratings, which further reduces the amount of ratings needed to reach a certain level of accuracy by $\sim3 \times$. Finally, using these methods, we were able to conduct a large-scale human study with reasonable cost (lowering cost from estimated \$90,000 USD to \$2160 USD) without sacrificing accuracy -- We evaluated 5 of the best performing methods for explaining vision neurons and discovered that Linear Explanations \cite{oikarinen2024linear} performed the best overall.
{\small
    \bibliographystyle{ieeenat_fullname}
    \bibliography{main}
}
\clearpage
\newpage

\onecolumn
\appendix
\section*{Appendix Overview}
\startcontents[sections]
\printcontents[sections]{l}{1}{\setcounter{tocdepth}{2}}

\newpage

\section{Discussion}

\subsection{Limitations}
\label{subsec:limit}
Due to time and budget constraints, our crowdsourced evaluation focuses on comparing descriptions on a few models/layers e.g. later layer neurons of ImageNet trained models. While we mostly find similar trends between the ResNet and ViT models, different description methods may have advantage on different types of neurons. For example, it is possible that Network Dissection based methods would perform better than they did on our current evaluation if we focus on lower layer neurons or on models trained on Places365 \cite{zhou2017places}, as the labels in the Broden dataset are more suitable for these tasks. Similarly, we think LLM based methods \cite{bai2025interpret, shaham2024multimodal} might perform better when describing neurons of a sparse autoencoder \cite{bricken2023monosemanticity} as these are more monosemantic and can be better described by highly activating inputs only.

Second, our crowdsourced evaluation relies on Amazon Mechanical Turk workers, who are not experts and often make errors in labeling. While we introduced principled measures to estimate the errors as well as mitigations for error, we cannot improve their domain knowledge, which means a crowdsourced evaluation might favor simpler descriptions over more complex concepts requiring domain knowledge. To compare more complex descriptions or neuron descriptions in a specific domain it may be necessary to recruit domain expert raters.

Finally, our evaluation is focused on evaluating \textit{input-based} neuron explanations that aim to explain the "input $\rightarrow$ neuron activation" function. Some recent work such as \cite{gandelsman2024interpreting, gur2025enhancing} instead focus on \textit{output-based} neuron explanations that aim to explain the "neuron activation $\rightarrow$ model output" relationship. Rigorously evaluating these \textit{output-based} explanations will require different methodology and is an interesting problem for future work.

\subsection{Broader Impact}
\label{subsec:broader_impact}
This paper is focused on better understanding neural networks via interpretability, and as such we expect it's impact to be largely positive as better understanding of neural networks can help us for example identify failure modes before deploying and enable better control of models. As our focus is in particular on rigorous evaluation of neuron explanations, this can help avoid interpretability illusions or users over-relying on unreliable explanations.

\subsection{Crowdsourced Study Design}
\label{sec:crowdsourced_study_design}

\begin{figure*}[h!]
    \centering
    \begin{subfigure}[b]{0.49\textwidth}
        \centering
        \includegraphics[width=\textwidth]{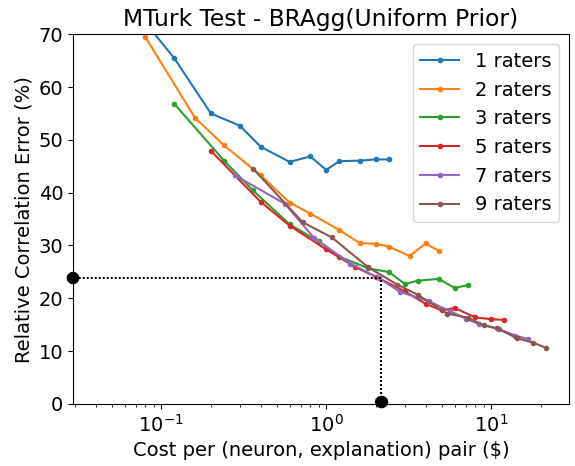}
        \caption{Aggregation with Method 3a: BRAgg - Uniform Prior.}
        \label{fig:n_raters_mturk_bayes_uniform}
    \end{subfigure}
    \hfill
    \begin{subfigure}[b]{0.49\textwidth}
        \centering
        \includegraphics[width=\textwidth]{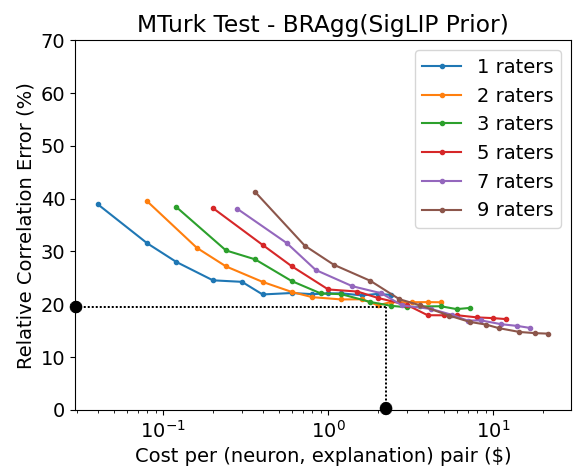}
        \caption{Aggregation with Method 3b: BRAgg - SigLIP prior.}
        \label{fig:n_raters_mturk_bayes_siglip}
    \end{subfigure}
    \caption{Comparing rating aggregation strategies in Setting 2 (Testing on MTurk) described in Sec~\ref{sec:validation}. The $x$-axis represents the cost of evaluations as number of inputs per neuron multiplied by number of raters per input, times cost per input. The lowest curve represents the optimal number of raters for a particular budget.}
    \label{fig:n_raters_mturk}
\end{figure*}

In this section we discuss how our methodology testing/validation settings described in Section \ref{sec:validation} can be used to help design the finer parameters of our study such as how many raters to use per input. In our current MTurk setup, the cost of obtaining a rating for a single image by a single rater is $\frac{\$0.06}{15} = \$0.004$ as we show 15 inputs per task. We can then plot the expected error as a function of evaluation cost by plotting $n_{inputs} \cdot n_{raters} \cdot \$0.004$ on the x-axis. 

In Figure \ref{fig:n_raters_mturk}, we plot the cost vs expected error rate for different numbers of raters we have per input. As we can see, the optimal number of raters per input depends on your budget, with smaller number of raters working better with smaller budgets, while for a larger budget it is optimal to use more raters per input.

Given our budget, we are aiming for a cost of around \$2.2 per (neuron, explanation pair), and we can see the optimal number of raters and expected error rate for this budget following the black lines on our plot. Since we wish to get good results using both aggregation methods, we can see 3 raters (\textcolor{Green}{green line}) is among the best for both methods, we choose to use 3 raters and 180 inputs per neuron for the crowd-sourced study. This gives us an expected correlation error of around $25\%$ with Bayes - Uniform Prior and around $20\%$ with Bayes - SigLIP prior based on the MTurk test.

\section{Ablation Studies}

\subsection{Choice of $\gamma$}
\label{app:gamma_choice}

\begin{figure}[h!]
    \centering
    \includegraphics[width=0.7\linewidth]{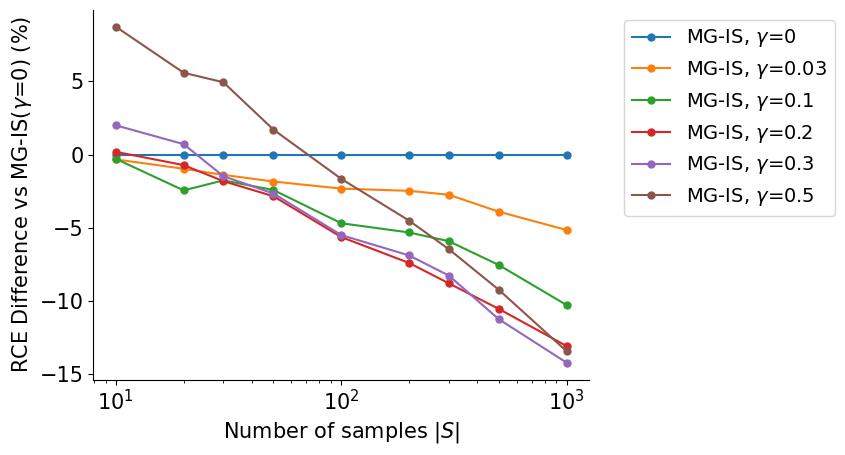}
    \caption{The Effect of the $\gamma$ hyperparameter on Importance sampling performance.}
    \label{fig:gamma_comparison}
\end{figure}

Figure \ref{fig:gamma_comparison} shows the effects of the hyperparameter $\gamma$ (Eq. \ref{eq:mg_is}) on importance sampling performance on our simulated setup with no label noise. For visual clarity, we report the difference compared to baseline with $\gamma=0$ instead of absolute numbers, so for example reducing RCE from $25\%$ to $22.5\%$ would show as $-10\%$ on this plot. Overall we can see our method is not very sensitive to this choice, but using a reasonable $\gamma$ does reduce error by around 10\%, with more relative difference on larger sample sizes. We use $\gamma=0.2$ as it performs the best on sample sizes around 200 that we use in our large study.

\subsection{Sensitivity to $\beta$ in BRAgg (Uniform Prior)}
\label{app:beta_ablation}

\begin{figure}[h!]
    \centering
    \includegraphics[width=0.55\linewidth]{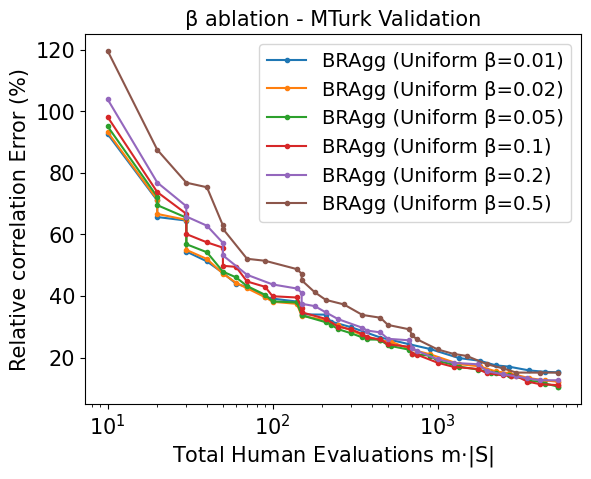}
    \caption{Comparing the error rate of using different $\beta$ values for BRAgg(Uniform prior).}
    \label{fig:uniform_prior_beta_comp}
\end{figure}

Our Rating Aggregation Method 3a Bayes - Uniform Sampling described in Section \ref{subsec:challenge_2} depends on the $\beta$ hyperparameter to select the uniform prior. In this section, we conducted a test on Setting 2 (MTurk test, using MG-IS sampling) comparing different values for $\beta$. As shown in Fig \ref{fig:uniform_prior_beta_comp}, the exact choice of $\beta$ has little effect on the correlation error, with values between 0.02 and 0.2 performing well. In our experiments we used $\beta=0.05$ which performs well overall. We think the small sensitivity is mostly because changing prior has a relatively linear effect on predicted $c_t$, and since correlation coefficient normalizes the $c_t$ the scale of $c_t$ does not change the correlation. Each datapoint in Figure \ref{fig:uniform_prior_beta_comp} is the average over 30 samples with different random seeds.

\newpage

\subsection{Activation-Guided Importance Sampling}
\label{app:act_guide_is}

\begin{figure}[h!]
    \centering
    \includegraphics[width=0.55\linewidth]{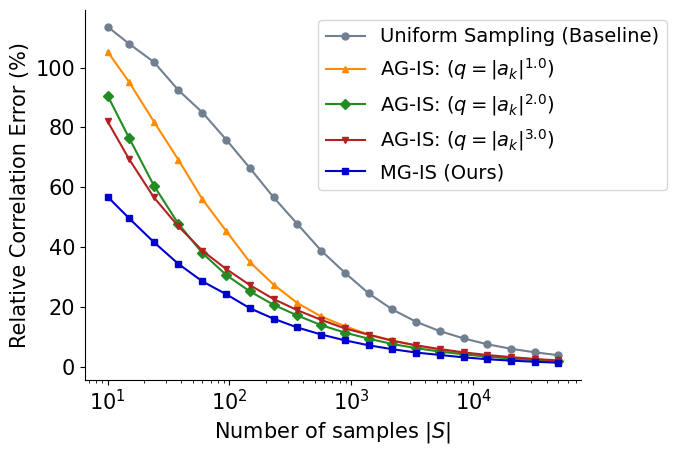}
    \caption{Comparing different sampling strategies on simulated data with no label noise. We can see importance sampling without model guidance can still perform quite well and deliver significant gains over uniform sampling.}
    \label{fig:sampling_simulation_act_guided}
\end{figure}

In addition to Model-Guided Importance Sampling (MG-IS), we also tested importance sampling guided purely by neuron activations, which we call activation guided importance sampling (AG-IS). This could be useful for cases where no cheap model is available to estimate concept presence. In particular we tested with importance sampling functions of the following form:

\begin{equation}
    q^{act}(x_i) = \frac{|[\bar{a}_{k}]_i|^{\alpha}}{\sum_{x_i \in \mathcal{D}} |[\bar{a}_{k}]_i|^{\alpha}}
\end{equation}

\begin{equation}
    q^{\text{AG-IS}}(x) = (1-\gamma) q^{act}(x) + \gamma p(x)
    \label{eq:ag_is}
\end{equation}

As shown in Figure \ref{fig:sampling_simulation_act_guided}, we can see very significant gains over uniform sampling, with the best performing variant being $\alpha=2$, but overall performance is not as good as our model-guided importance sampling. We used $\gamma = 0.2$ for these experiments.

\subsection{Error Model for Rating Aggregation}
\label{app:error_model_comparison}

For our main results, our Bayes modeling uses a uniform error model, i.e. we assume all have concepts and inputs share the same error rate $\eta$ as described in Section \ref{subsec:challenge_2}. However, this is not always a realistic assumption, as some inputs and concept are harder/more ambiguous, leading to higher error rates. To account for this, we also experimented using an alternative error model where the error rate $\eta_{ti}$ is different for each pair of input $i$ and concept $t$.

In this model $\mathbb{P}(r^j_{ti} = c_{ti}^{*}) = 1-\eta_{ti}$, where $\eta_{ti} \sim \text{Beta}(0.75, 2.5)$. The hyperparameters $\alpha = 0.75$ and $\beta = 2.5$ were selected to maximize the probability of the MTurk user ratings we observed in Setting 2.

To do Bayesian inference with this model, we do inference on both $c_{ti}^*$ and $\eta_{ti}$ and report $[\concept]_i = \int_{\eta_{ti}} \mathbb{P}(C_{ti}, \eta_{ti} | R_{ti}) d \eta_{ti}$.

Applying the Bayes rule $\mathbb{P}(C_{ti}, \eta_{ti}|R_{ti}) = \mathbb{P}(R_{ti}|C_{ti}, \eta_{ti})\mathbb{P}(C_{ti}, \eta_{ti})/\mathbb{P}(R_{ti})$ we get:

\begin{equation}
    [\concept]_i =  \frac{ \int_{\eta_{ti}} \mathbb{P}(R_{ti} | C_{ti}, \eta_{ti})\cdot \mathbb{P}(C_{ti}) \mathbb{P}(\eta_{ti}) d \eta_{ti}}{\int_{\eta_{ti}} \mathbb{P}(R_{ti} | C_{ti}, \eta_{ti}) \cdot \mathbb{P}(C_{ti})\mathbb{P}(\eta_{ti}) + \mathbb{P}(R_{ti} | \lnot C_{ti}, \eta_{ti})\mathbb{P}(\lnot C_{ti})\mathbb{P}(\eta_{ti}) d \eta_{ti}} 
\end{equation}

Since $\mathbb{P}(\eta_{ti})$ and $\mathbb{P}(\_{ti})$ are independent. In practice calculating exact solution for the integral is difficult, so we instead approximate it with 100 discrete bins. \newline

\textbf{Likelihood:} $\mathbb{P}(R_{ti} | C_{ti}, \eta_{ti})$. 

With $\alpha_{ti} = \sum_{j=1}^m r_{ti}^j$, we obtain the likelihood in below equations:
\begin{equation}
    \mathbb{P}(R_{ti} | C_{ti}, \eta_{ti}) = (1-\eta_{ti})^{\alpha_{ti}}(\eta_{ti})^{(m-\alpha_{ti})}
\end{equation}
\begin{equation}
    \mathbb{P}(R_{ti} | \lnot C_{ti}, \eta_{ti}) = (\eta_{ti})^{\alpha_{ti}}(1-\eta_{ti})^{(m-\alpha_{ti})}
\end{equation}

\textbf{Priors:}
For $\mathbb{P}(C_{ti})$ we use the same priors as in Section \ref{subsec:challenge_2} (uniform or SigLIP prior). For $\mathbb{P}(\eta_{ti})$, we use the Beta distribution:
\begin{equation}
    \mathbb{P}(\eta_{ti}) \propto \eta_{ti}^{\alpha} \cdot (1-\eta_{ti})^{\beta}
\end{equation}

The hyperparameters $\alpha = 0.75$ and $\beta = 2.5$ were selected to maximize the probability of the MTurk user ratings we observed.

\paragraph{Results}

\begin{figure}[h!]
    \centering
    \includegraphics[width=0.5\linewidth]{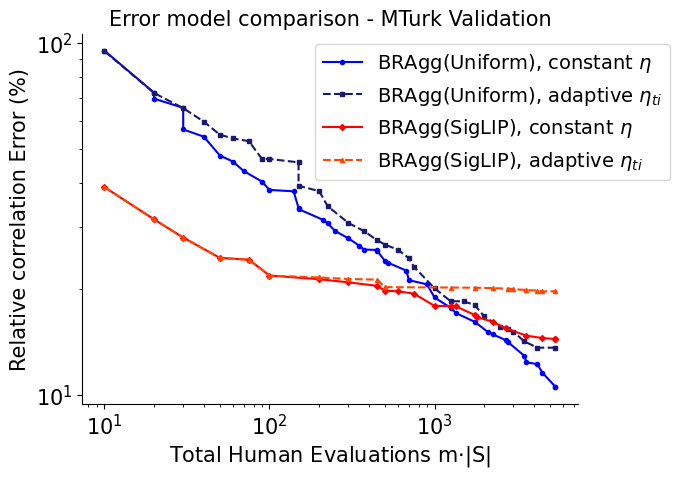}
    \caption{Comparing different error models for rating aggregation. Note log-scale on both axes.}
    \label{fig:error_model_comparison}
\end{figure}

As shown in Figure \ref{fig:error_model_comparison}, the performance between different error models is similar, but overall the simplified uniform error model (solid lines) leads to slightly better results, so we use the simple error model for our main results. We are not fully certain why the simple model performs better, but one possibility is that even 9 ratings per user is not enough for the Bayes model to accurately infer per-input error rates.

\subsection{SigLIP Only Evaluation Accuracy}
\label{app:siglip_accuracy}

\begin{figure}
    \centering
    \includegraphics[width=0.6\linewidth]{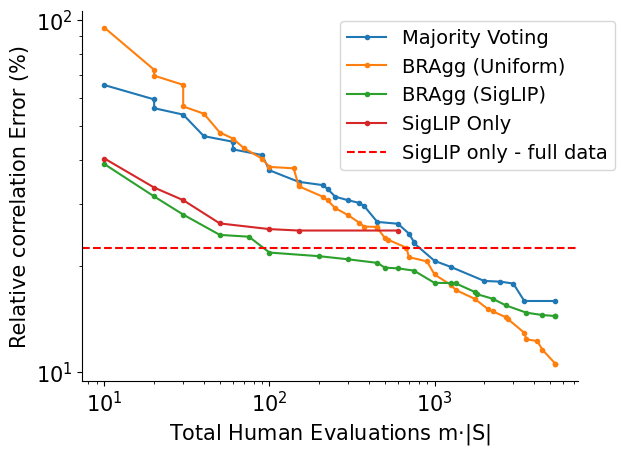}
    \caption{Comparison the error of different rating aggregation strategies and only relying on SigLIP instead of conducting a human study. Using our Setting 2 results from MTurk with MG-IS sampling.}
    \label{fig:siglip_only_comparison}
\end{figure}

Figure \ref{fig:siglip_only_comparison} shows comparison of different rating aggregation strategies vs a pure automated evaluation only relying on SigLIP scores. We can see that the optimal rating strategy depends on your budget. A fully automated evaluation provides similar accuracy as a hybrid evaluation (BRAgg(SigLIP)) with ~100 samples per (neuron, explanation) pair, or a pure human evaluation with ~800 samples per neuron. This implies that the optimal study strategy depends on your budget: If you cannot afford more than 100 human evaluations per neuron, it is best to rely on automated evaluation. For budgets between 100-1000 samples per neuron, a hybrid evaluation (BRAgg(SigLIP)) performs the best. For sufficiently large budget, i.e. >1000 evaluations per neuron, pure human evaluation (BRAgg(Uniform)) performs the best.

\subsection{Compositional/Complex Explanations}
\label{app:complex_explanations}

A more complex explanation is typically more accurate, but it is harder to understand, and more expensive to evaluate as it requires labels for each of the concepts involved in the explanation. We use the length $l$ to indicate explanation complexity, where $l$ is the number of unique concepts in the explanation, e.g. "cat" OR "dog" would have lenght 2.

\begin{table*}[h!]
\centering
\scalebox{0.83}{
\begin{tabular}{@{}lcccccc@{}}
\toprule
\multicolumn{1}{c}{Complex Exp. ($l$>1)} & \begin{tabular}[c]{@{}c@{}}Comp Exp~\cite{mu2021compositional}\\ l=3\end{tabular} & \begin{tabular}[c]{@{}c@{}}INVERT~\cite{bykov2023labeling}\\ l=3\end{tabular} & \begin{tabular}[c]{@{}c@{}}CCE~\cite{la2023towards},\\ l= 5\end{tabular} & \begin{tabular}[c]{@{}c@{}}CCE~\cite{la2023towards},\\ l=15\end{tabular} & \begin{tabular}[c]{@{}c@{}}LE(label)~\cite{oikarinen2024linear}\\ l=4.37/1.97\end{tabular} & \begin{tabular}[c]{@{}c@{}}LE(SigLIP)~\cite{oikarinen2024linear} \\ l=4.66/1.82\end{tabular} \\ \midrule
\begin{tabular}[c]{@{}l@{}}RN-50 \\ (Layer4)\end{tabular} & \begin{tabular}[c]{@{}c@{}}0.1399\\ $\pm$ 0.002\end{tabular} & \begin{tabular}[c]{@{}c@{}}0.2341\\ $\pm$ 0.002\end{tabular} & \begin{tabular}[c]{@{}c@{}}0.0993 \\ $\pm$ 0.002\end{tabular} & \begin{tabular}[c]{@{}c@{}}0.1510 \\ $\pm$ 0.003\end{tabular} & \begin{tabular}[c]{@{}c@{}}\underline{0.2924} \\ \underline{$\pm$ 0.002}\end{tabular} & \textbf{\begin{tabular}[c]{@{}c@{}}0.3772\\ $\pm$ 0.002\end{tabular}} \\ \midrule
\begin{tabular}[c]{@{}l@{}}ViT-B-16\\ (Layer11 MLP)\end{tabular} & \begin{tabular}[c]{@{}c@{}}0.0468\\ $\pm$ 0.002\end{tabular} & \begin{tabular}[c]{@{}c@{}}0.1101\\ $\pm$ 0.003\end{tabular} & \begin{tabular}[c]{@{}c@{}}0.0534\\ $\pm$ 0.003\end{tabular} & \begin{tabular}[c]{@{}c@{}}0.0570 \\ $\pm$ 0.006\end{tabular} & \begin{tabular}[c]{@{}c@{}}\underline{0.3243}\\ \underline{$\pm$ 0.005}\end{tabular} & \textbf{\begin{tabular}[c]{@{}c@{}}0.3489\\ $\pm$ 0.005\end{tabular}} \\ \bottomrule \\
\end{tabular}}
\caption{SigLIP based simulation with correlation scoring comparing complex explanations ($l>1$). We can see Linear Explanation (LE) overall performs the best.}
\label{tab:siglip_sim_complex}
\end{table*}

For fairness we split the explanations into two groups, simple explanations where $l=1$ and complex explanations with $l>1$. Table \ref{tab:siglip_sim_len1} in main text shows comparison of different simple explanations, while Table \ref{tab:siglip_sim_complex} compares complex explanation methods. For complex explanations, we can see Linear Explanations reaches the highest correlations with most methods having higher scores than their simple counterpart, highlighting the value of added complexity.

\subsection{Model Ablation}

In our main results we used the SigLIP \cite{zhai2023sigmoid} ViT-L-16-384 model both as the guide for importance sampling (MG-IS), and to provide the prior for our Bayes rating aggregation. For the Automated Evaluation described in Section \ref{subsec:automated_eval} we used the SigLIP ViT-SO400M-14-384 model.

The purpose of this section is to analyze the sensitivity of our results to the model choice. In particular, we test replacing the SigLIP ViT-L-16-384 model for sampling and rating aggregation with the following models:
\begin{enumerate}
    \item CLIP \cite{radford2021learning} ViT-L14-336, which is an older model with slightly weaker benchmark performance.
    \item SigLIP2 \cite{tschannen2025siglip} ViT-gopt-16-384, which is a bigger and more recent model with slightly bettet benchmark scores.
\end{enumerate}

For all models we used the weights and implementation from the open\_clip package \cite{ilharco_gabriel_2021_5143773}. 

\begin{figure}[h!]
    \centering
    \includegraphics[width=0.5\linewidth]{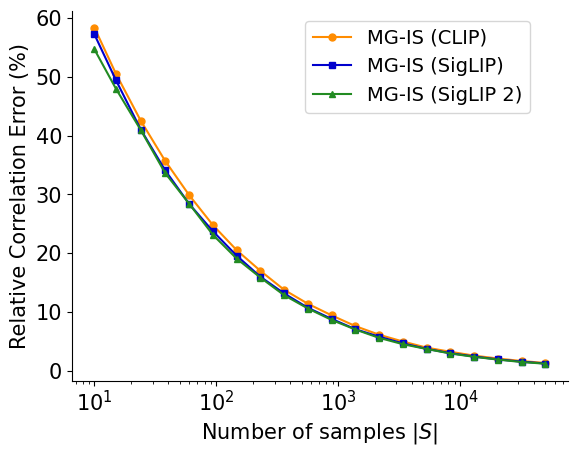}
    \caption{Comparing different models for guiding sampling in our simulated test (Setting 1). We can see stronger models improve sampling but the differences are small.}
    \label{fig:sampling_model_ablation}
\end{figure}

\paragraph{Results - Sampling Model:} Figure \ref{fig:sampling_model_ablation} shows a comparison of different models for guiding sampling. We can see better models improve results but the differences are quite small. On average using CLIP instead of SigLIP leads to around 5 \% (relative) increase in error with the same number of samples, while using SigLIP 2 leads to around 2 \% (relative) reduction in error.

\begin{figure}[h!]
    \centering
    \includegraphics[width=0.5\linewidth]{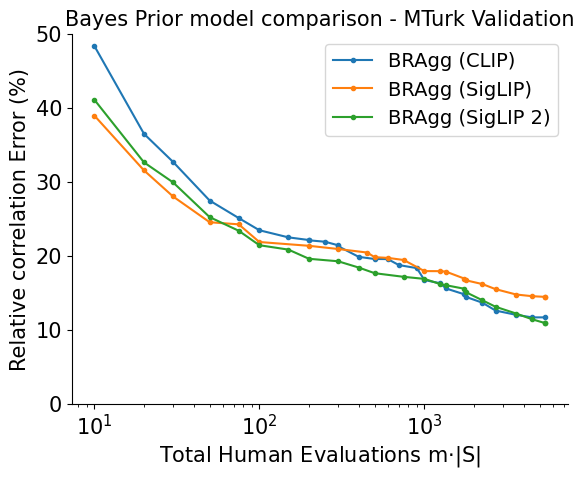}
    \caption{Comparing different choices for the Bayes prior. We can see SigLIP2 performs the best overall.}
    \label{fig:bayes_prior_model_ablation}
\end{figure}

\paragraph{Results - Model for Bayes Prior:} When changing the model used for creating the prior in BRAgg, we observe similar trends, with larger models generally performing better than smaller ones. Figure \ref{fig:bayes_prior_model_ablation} shows the results of using different models as the Bayes prior 
We notice a somewhat bigger performance difference in this setting, with for SigLIP 2 reducing RCE at (3 raters per input and 200 inputs) from 19.70\%(of SigLIP) to 17.95\% for around $10\%$ relative reduction in error. Interestingly, using CLIP as the prior performs worse with small sample sizes but beats SigLIP with large evaluation budget. Inspired by these results, we also analyze our main experiment results using SigLIP 2 instead of SigLIP as the Bayes prior and report the results in Table \ref{tab:siglip2_large_scale}. Overall we can see the results are very similar to our results using SigLIP as the prior reported in Table \ref{tab:mturk_res}. The only difference we notice is the correlation scores are slightly higher for all methods on ViT-B-16 neurons. This highlights our conclusions regarding the relative performance of different explanation methods are not sensitive to model choice.

\begin{table}[]
\centering
\begin{tabular}{lcccccc}
\hline
\textbf{Model} & \textbf{Aggregation} & \textbf{CLIP-Dissect} & \textbf{INVERT l=1} & \textbf{DnD} & \textbf{MAIA} & \textbf{LE(SigLIP) l=1} \\ \hline
\textbf{\begin{tabular}[c]{@{}l@{}}ResNet-50\\ (Layer4)\end{tabular}} & \begin{tabular}[c]{@{}c@{}}BRAgg \\ (SigLIP 2 Prior)\end{tabular} & \begin{tabular}[c]{@{}c@{}}0.1392 \\ $\pm$ 0.008\end{tabular} & \begin{tabular}[c]{@{}c@{}} \underline{0.1490} \\ \underline{$\pm$ 0.005} \end{tabular} & \begin{tabular}[c]{@{}c@{}}0.1242 \\ $\pm$ 0.008\end{tabular} & \begin{tabular}[c]{@{}c@{}}0.1072 \\ $\pm$ 0.008\end{tabular} & \textbf{\begin{tabular}[c]{@{}c@{}}0.1786 \\ $\pm$ 0.009\end{tabular}} \\ \hline

\textbf{\begin{tabular}[c]{@{}l@{}}ViT-B-16\\ (Layer11 mlp)\end{tabular}} & \begin{tabular}[c]{@{}c@{}}BRAgg \\ (SigLIP 2 Prior)\end{tabular} & \begin{tabular}[c]{@{}c@{}}\underline{0.1714} \\ \underline{$\pm$ 0.019}\end{tabular} & \begin{tabular}[c]{@{}c@{}}0.1250 \\ $\pm$ 0.021\end{tabular} & \begin{tabular}[c]{@{}c@{}}0.1039\\ $\pm$ 0.016\end{tabular} & \begin{tabular}[c]{@{}c@{}}0.1355 \\ $\pm$ 0.019\end{tabular} & \textbf{\begin{tabular}[c]{@{}c@{}}0.2591 \\ $\pm$ 0.023\end{tabular}} \\ \hline
\textbf{Average} & \begin{tabular}[c]{@{}c@{}}BRAgg \\ (SigLIP 2 Prior)\end{tabular} & \underline{0.1553} & 0.1370 & 0.1141 & 0.1201 & \textbf{0.2189} \\ \hline
\end{tabular}
\caption{Large scale crowdsourced study results using SigLIP 2 as the prior for BRAgg.}
\label{tab:siglip2_large_scale}
\end{table}
\clearpage

\begin{figure*}[h!]
    \centering
    \includegraphics[width=0.99\linewidth]{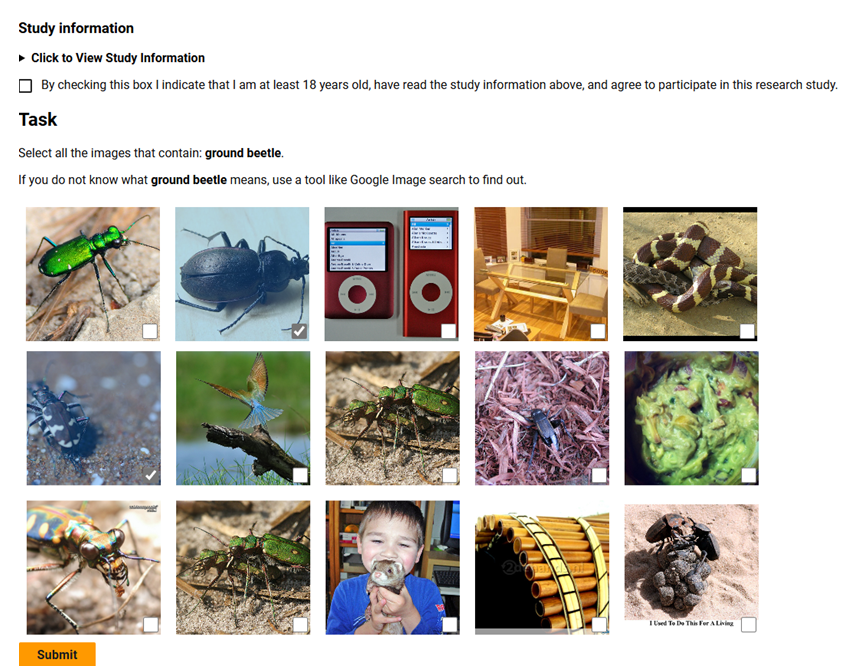}
    \caption{Our user study interface.}
    \label{fig:user_interface}
\end{figure*}

\section{Additional Details}
\label{sec:extra_details}

\subsection{MTurk Experimental Details:}
\label{app:mturk_details}

Figure \ref{fig:user_interface} showcases the full user interface displayed to the raters. We selected raters based in the US with over 10,000 tasks approved and $>98\%$ task approval rate. Each rater was paid \$0.05 per task(and we paid another \$0.01 per task in fees to MTurk), which takes around 15 seconds to complete based on our testing, for an estimated \$12/hr earnings.

\subsection{Detailed MTurk results}
\label{app:detailed_mturk_results}


Table \ref{tab:mturk_res} shows more detailed results of our crowdsourced evaluation, including multiple rating aggregation methods. Comparing different rating aggregation methods, we can see that the correlation scores themselves vary a lot depending on the aggregation method used, but the order between explanations methods is quite consistent, for example Linear Explanations \cite{oikarinen2024linear} score the highest regardless of aggregation method used.


\begin{table}[]
\centering
\begin{tabular}{@{}lcccccc@{}}
\toprule
\textbf{Model} & \textbf{Aggregation} & \textbf{CLIP-Dissect\cite{oikarinen2023clip}} & \textbf{INVERT l=1\cite{bykov2023labeling}} & \textbf{DnD}\cite{bai2025interpret} & \textbf{MAIA}\cite{shaham2024multimodal} & \textbf{LE(SigLIP) l=1}\cite{oikarinen2024linear} \\ \midrule

\textbf{\begin{tabular}[c]{@{}l@{}}ResNet-50\\ (Layer4)\end{tabular}} & Majority & \begin{tabular}[c]{@{}c@{}}0.0978 \\ $\pm$ 0.010\end{tabular} & \begin{tabular}[c]{@{}c@{}}\underline{0.0981} \\ \underline{$\pm$ 0.010}\end{tabular} & \begin{tabular}[c]{@{}c@{}}0.0978 \\ $\pm$ 0.010\end{tabular} & \begin{tabular}[c]{@{}c@{}}0.0834 \\ $\pm$ 0.010\end{tabular} & \textbf{\begin{tabular}[c]{@{}c@{}}0.1216 \\ $\pm$ 0.011\end{tabular}} \\

\cmidrule(l){2-7} 
 & \begin{tabular}[c]{@{}c@{}}BRAgg \\ (Uniform Prior)\end{tabular} & \begin{tabular}[c]{@{}c@{}} \underline{0.1017} \\ \underline{$\pm$ 0.009}\end{tabular} & \begin{tabular}[c]{@{}c@{}}0.0977 \\ $\pm$ 0.010\end{tabular} & \begin{tabular}[c]{@{}c@{}}0.1011 \\ $\pm$ 0.008\end{tabular} & \begin{tabular}[c]{@{}c@{}}0.0876 \\ $\pm$ 0.010\end{tabular} & \textbf{\begin{tabular}[c]{@{}c@{}}0.1261 \\ $\pm$ 0.011\end{tabular}} \\ 
 
 \cmidrule(l){2-7}
 & \begin{tabular}[c]{@{}c@{}}BRAgg \\ (SigLIP Prior)\end{tabular} & \begin{tabular}[c]{@{}c@{}}0.1356 \\ $\pm$ 0.008\end{tabular} & \begin{tabular}[c]{@{}c@{}}\underline{0.1472} \\ \underline{$\pm$ 0.005}\end{tabular} & \begin{tabular}[c]{@{}c@{}}0.1179 \\ $\pm$ 0.007\end{tabular} & \begin{tabular}[c]{@{}c@{}}0.0992 \\ $\pm$ 0.008\end{tabular} & \textbf{\begin{tabular}[c]{@{}c@{}}0.1803 \\ $\pm$ 0.008\end{tabular}} \\ 
 
 \midrule
\textbf{\begin{tabular}[c]{@{}l@{}}ViT-B-16\\ (Layer11 mlp)\end{tabular}} & Majority vote & \begin{tabular}[c]{@{}c@{}}\underline{0.1035} \\ \underline{$\pm$ 0.016}\end{tabular} & \begin{tabular}[c]{@{}c@{}}0.0754 \\ $\pm$ 0.016\end{tabular} & \begin{tabular}[c]{@{}c@{}}0.0825 \\ $\pm$ 0.017\end{tabular} & \begin{tabular}[c]{@{}c@{}}0.0723 \\ $\pm$ 0.013\end{tabular} & \textbf{\begin{tabular}[c]{@{}c@{}}0.1208 \\ $\pm$ 0.015\end{tabular}} \\ 

\cmidrule(l){2-7} 
 & \begin{tabular}[c]{@{}c@{}}BRAgg \\ (Uniform Prior)\end{tabular} & \begin{tabular}[c]{@{}c@{}}\underline{0.0934} \\ \underline{$\pm$ 0.014}\end{tabular} & \begin{tabular}[c]{@{}c@{}}0.0637 \\ $\pm$ 0.014\end{tabular} & \begin{tabular}[c]{@{}c@{}}0.0689 \\ $\pm$ 0.013\end{tabular} & \begin{tabular}[c]{@{}c@{}}0.0555 \\ $\pm$ 0.012\end{tabular} & \textbf{\begin{tabular}[c]{@{}c@{}}0.1018 \\ $\pm$ 0.012\end{tabular}} \\ 
 
 \cmidrule(l){2-7} 
 & \begin{tabular}[c]{@{}c@{}}BRAgg \\ (SigLIP Prior)\end{tabular} & \begin{tabular}[c]{@{}c@{}}\underline{0.1568} \\ \underline{$\pm$ 0.017}\end{tabular} & \begin{tabular}[c]{@{}c@{}}0.1167 \\ $\pm$ 0.020\end{tabular} & \begin{tabular}[c]{@{}c@{}}0.0904 \\ $\pm$ 0.015\end{tabular} & \begin{tabular}[c]{@{}c@{}}0.1257 \\ $\pm$ 0.017\end{tabular} & \textbf{\begin{tabular}[c]{@{}c@{}}0.2450 \\ $\pm$ 0.021\end{tabular}} \\ \midrule
\textbf{Average} & \begin{tabular}[c]{@{}c@{}}BRAgg \\ (SigLIP Prior)\end{tabular} & \underline{0.1462} & 0.132 & 0.1042 & 0.1125 & \textbf{0.2127} \\ \cmidrule(l){2-7} 
 & All & \underline{0.1148} & 0.0998 & 0.0931 & 0.0873 & \textbf{0.1493} \\ \bottomrule
\end{tabular}
 \caption{Detailed results of our large scale crowdsourced study. For each model we tested 100 random neurons, and we report the average correlation score of the explanations produced by each method, as well as the standard error of the mean. Comparing different rating aggregation methods, we can see that the correlation scores themselves vary a lot depending on the aggregation method used, but the order between explanations methods is quite consistent.}
 \label{tab:mturk_res}
\end{table}

\paragraph{Statistical Significance}

\begin{table}[h!]
\centering
\begin{tabular}{@{}llcccc@{}}
\toprule
 &  & \multicolumn{4}{c}{Hypothesis} \\ \midrule
Model & Aggregation & \multicolumn{1}{l}{LE \textgreater CLIP-Dissect} & \multicolumn{1}{l}{LE \textgreater INVERT} & \multicolumn{1}{l}{LE \textgreater DnD} & \multicolumn{1}{l}{LE \textgreater MAIA} \\ \midrule
RN-50 & BRAgg(Uniform) & 0.0601 & \textbf{0.0079} & \textbf{0.0198} & \textbf{0.0010} \\
 & BRAgg(SigLIP) & \textbf{0.0020} & \textbf{0.0098} & \textbf{0.0000} & \textbf{0.0000} \\
ViT-B-16 & BRAgg(Uniform) & 0.4295 & \textbf{0.0037} & \textbf{0.0431} & \textbf{0.0062} \\
 & BRAgg(SigLIP) & \textbf{0.0039} & \textbf{0.0001} & \textbf{0.0000} & \textbf{0.0001} \\ \bottomrule
\end{tabular}
\caption{Statistical significance of our results according to two-sample students t-test. We report the p-values of different hypothesis, with statistically significant p-values in bold.}
\label{tab:statistical_significance}
\end{table}

Table \ref{tab:statistical_significance} reports the statistical significance of our results using the two-sample Student's t-test. We can see we have statistically significant evidence that Linear Explanations performs better than all the other methods, with the exception of CLIP-Dissect when using BRAgg with uniform prior, and all methods with the SigLIP prior.
 
\begin{figure}[h]
    \centering
    \begin{minipage}{0.48\textwidth}
        \centering
        \includegraphics[width=\textwidth]{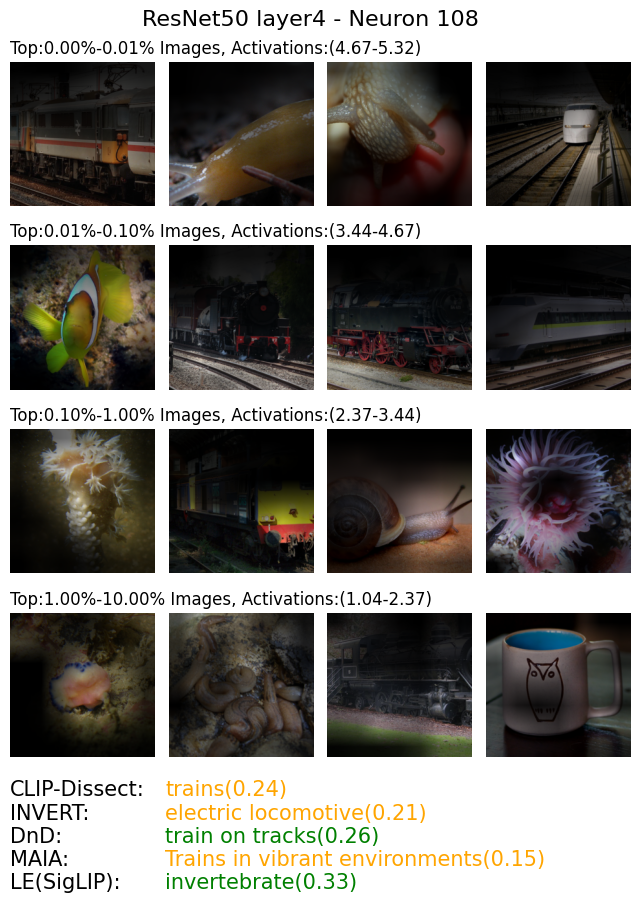}
    \end{minipage}
    \hfill
    \begin{minipage}{0.48\textwidth}
        \centering
        \includegraphics[width=\textwidth]{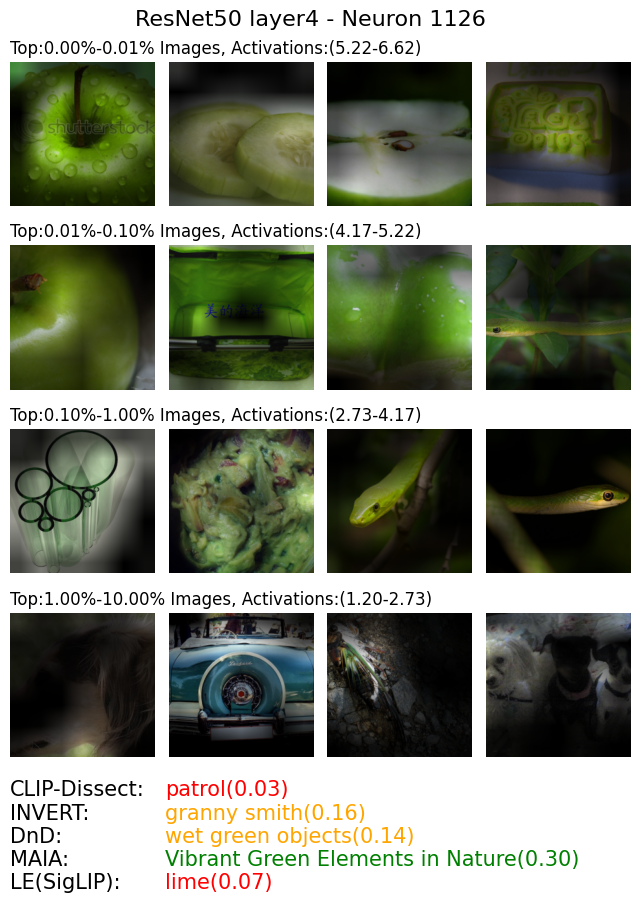}
    \end{minipage}
    \caption{Visualization of example neurons, their descriptions and correlations scores from our crowdsourced evaluation (BRAgg with SigLIP prior). We have colored the descriptions based on the correlation score.}
    \label{fig:example_ratings_rn50_1}
\end{figure}

\begin{figure}[h]
    \centering
    \begin{minipage}{0.48\textwidth}
        \centering
        \includegraphics[width=\textwidth]{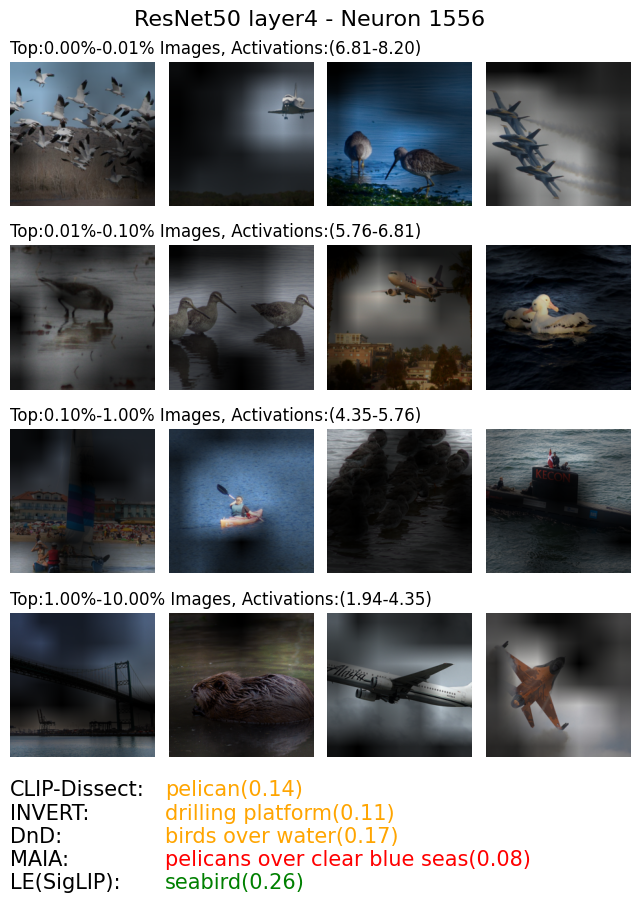}
    \end{minipage}
    \hfill
    \begin{minipage}{0.48\textwidth}
        \centering
        \includegraphics[width=\textwidth]{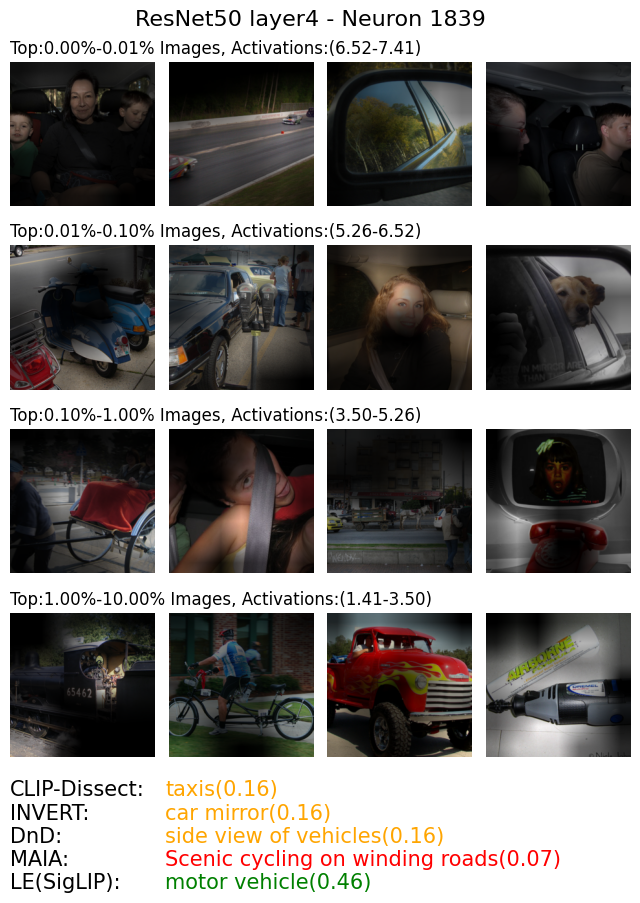}
    \end{minipage}
    \caption{Visualization of example neurons, their descriptions and correlations scores from our crowdsourced evaluation (BRAgg with SigLIP prior). We have colored the descriptions based on the correlation score.}
    \label{fig:example_ratings_rn50_2}
\end{figure}

\begin{figure}[h]
    \centering
    \begin{minipage}{0.48\textwidth}
        \centering
        \includegraphics[width=\textwidth]{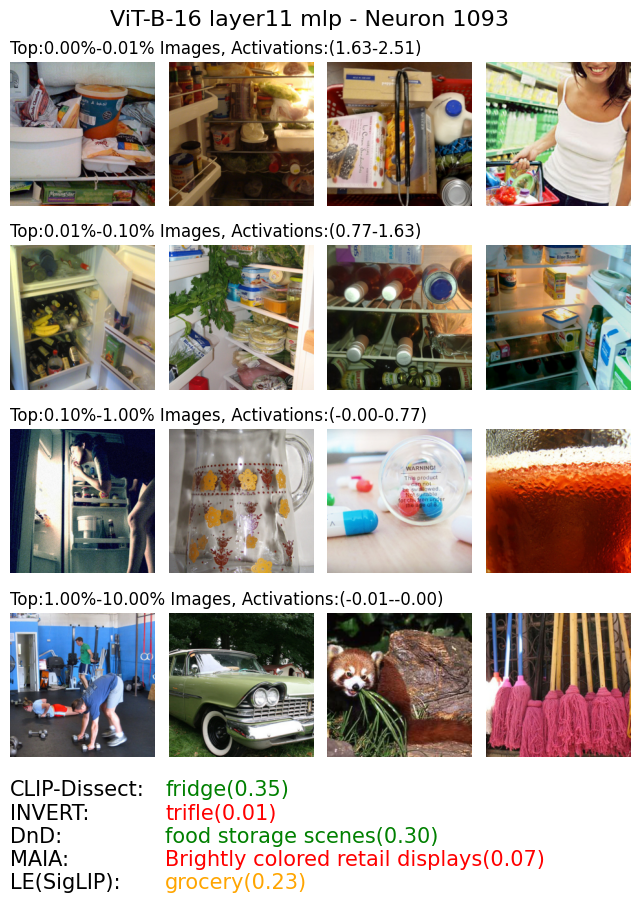}
    \end{minipage}
    \hfill
    \begin{minipage}{0.48\textwidth}
        \centering
        \includegraphics[width=\textwidth]{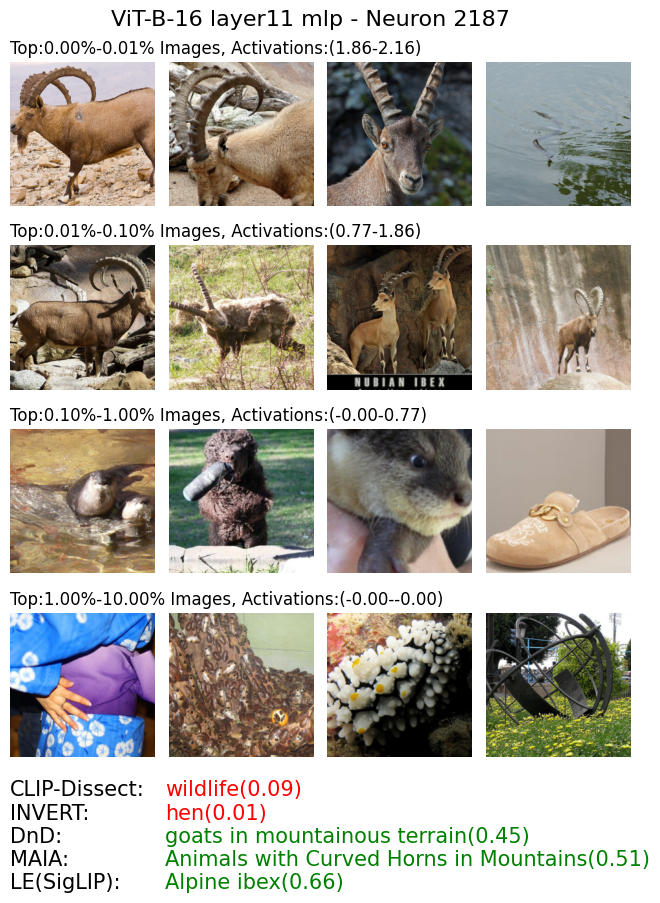}
    \end{minipage}
    \caption{Visualization of example neurons, their descriptions and correlations scores from our crowdsourced evaluation (BRAgg with SigLIP prior). We have colored the descriptions based on the correlation score.}
    \label{fig:example_ratings_vit}
\end{figure}

\paragraph{Example Ratings} Figures \ref{fig:example_ratings_rn50_1}, \ref{fig:example_ratings_rn50_2} and \ref{fig:example_ratings_vit} showcase example neurons and the descriptions assigned by different explanation methods, as well as the correlation scores estimated by our crowd-sourced study for those explanations. We colored explanations based on the estimated correlation coefficient, with Green: $\rho > 0.25$, Yellow: $ 0.25 \geq \rho > 0.10$ and Red: $ 0.10 \geq \rho$. 

We can see sometimes generative methods like MAIA \cite{shaham2024multimodal} produce the best explanations, for example "Vibrant Green Elements in Nature" in Figure \ref{fig:example_ratings_rn50_1}(Neuron 1126), but other times are too specific or fail to find the right concept \ref{fig:example_ratings_rn50_2}. 

Overall we did not observe very high correlation scores for any neurons in ResNet50, likely due to them being more densely activated and polysemantic, see for example neuron 108 (Fig. \ref{fig:example_ratings_rn50_1} that activates for both trains and invertebrates. On the other hand, for ViT-B-16 we observed several neurons that were extremely interpretable (correlation scores >0.5), such as Neuron 2187 (Fig. \ref{fig:example_ratings_vit}) that seems to only activate on Alpine ibex/Mountain goats.

\clearpage

\subsection{Experimental Details}
\label{sec:expt_details}

Our evaluation focused on two models/layers:
\begin{enumerate}[leftmargin=2em]
    \item Layer4 (end of residual block 4) of ResNet-50 trained on ImageNet. For simulation we use the neuron activation after average pooling, giving scalar activations, but for methods designed for 2d-activations, activations before average pooling are given as input.
    \item MLP-activations in Layer 11 of ViT-B-16 \cite{dosovitskiy2020image} Encoder trained on ImageNet. We focus only on the activations of the CLS-token as this is the last layer and other tokens do not affect the prediction.
\end{enumerate}

We use the full ImageNet validation dataset as $\mathcal{D}$ for the human study for all methods, and for generating explanations unless the method requires a specific dataset for explanation generation such as Broden \cite{netdissect2017}.

For all methods using SigLIP guidance we use the SigLIP-SO400M-14-386 model.


\subsubsection{Baseline Implementation Details}


For practical purposes we made slight modifications to some baseline methods. The changes are discussed in detail below:

\textbf{DnD} \cite{bai2025interpret}: The original implementation uses GPT-3.5 Turbo through the OpenAI API. Given the high cost of using the API and that recent open-source LLMs have strong performance compared to even closed-source LLMs like GPT-3.5, we replace GPT-3.5 Turbo with Llama 3.1-8B-Instruct \cite{llama3}. DnD \cite{bai2025interpret} showed that the older Llama 2 model was already better than GPT-3.5 Turbo and on-par with GPT-4 Turbo for neuron description, so our choice of Llama 3.1 does not degrade DnD's performance w.r.t.\ GPT-3.5 Turbo. 

\textbf{MAIA} \cite{shaham2024multimodal}: Compared to the original implementation, we replace GPT4-vision-preview with the newer GPT4o-2024-08-06 \cite{gpt4o} since it has lower API costs and better performance. The method is still quite expensive, costing us approximately \$65 and \$116 to generate descriptions for 100 randomly selected neurons of ResNet-50 Layer4 and ViT-B-16 Layer11 MLP respectively. Note that this cost also includes repeating the experiment for $\sim$10 and $\sim$20 neurons from ResNet and ViT respectively which did not yield any neuron description in the first run. Initially, we also tried open-source LLMs with support for vision inputs (\textit{i.e.}\ VLMs) like Llama-3.2-11B-Vision-Instruct \cite{llama3}, Llava-OneVision-Qwen2-7B-ov-hf \cite{llavaonevision}, and CogVLM2-Llama3-Chat-19B \cite{hong2024cogvlm2}. However, these VLMs do not work well with MAIA since they fail to generate executable code, forget the image tokens and focus only on the last few text tokens if given a long prompt, and allow only one image input at a time. This is likely because these VLMs are geared towards visual question answering and do not possess the more generalized capabilities of GPT4/4o.

\textbf{Methods relying on 2d activations:} Many methods are designed to explain entire channels of CNNs with 2d activations \cite{netdissect2017, mu2021compositional, hernandez2022natural, la2023towards}. For ResNet-50 layer4 we fed the pre-avg pool activations to these methods for proper 2d input. However, for ViT-B-16 last layer only the CLS-token activations affect the output, and as such we are explaining neurons with scalar activations. In this case, we broadcasted the scalar activations into a 2d-tensor with the same value in all spatial locations. However this is not the intended way to use these methods and may partially explain poorer performance of some methods on ViT-neurons as observed in Tables \ref{tab:siglip_sim_len1} and \ref{tab:siglip_sim_complex}.

\textbf{CCE} \cite{la2023towards}: For Clustered Compositional Explanations, we tested two different versions: the $l=15$ version corresponds to the default version with length 3 explanations for each of the 5 activation clusters. For the $l=5$ version we used explanation length=1 with 5 clusters of activations. We also used the implementation of \cite{la2023towards} to reproduce the results of Compositional Explanations \cite{mu2021compositional} by setting explanation length=3 and number of clusters=1.

\subsubsection{Subset of Neurons}

For most methods in Table \ref{tab:siglip_sim_len1} and \ref{tab:siglip_sim_complex}, we report the average correlation scores across all 2048 neurons for ResNet-50 layer4 and 3072 neurons for ViT-B-16 layer11 mlp. However for certain methods due to high computational and/or API cost we were only able to explain a subset of these neurons and report the average score of these subsets in Tables \ref{tab:siglip_sim_len1} and \ref{tab:siglip_sim_complex}. We report the results for a subset of neurons for the following methods:
\begin{itemize}
    \item MAIA~\cite{shaham2024multimodal}: Randomly selected subset of 100 neurons each for both RN50 and ViT-B-16.
    \item CCE~\cite{la2023towards} $l=5$: For ViT-B-16 we used a subset of 1420 neurons. RN50 evaluated on all neurons.
    \item CCE~\cite{la2023towards} $l=15$: RN50: subset of 984 neurons. ViT-B-16: subset of 422 neurons.
\end{itemize}
All other methods were evaluated on all neurons of each layer.

\subsection{Automated Evaluation Details}
\label{app:automated_eval_details}

For our automated evaluation(Sec. \ref{subsec:automated_eval}), we use Simulation with Correlation Scoring as described by \cite{oikarinen2024linear}. This evaluation was originally proposed for language model neurons by \cite{bills2023language}.

The basic idea of simulation evaluation to use the \textit{explanation} to predict neuron activations on unseen inputs. With correlation scoring we then evaluate the correlation coefficient $\rho$ between the predicted activations $s$ and actual neuron activations $a_k$ on the entire test split of 10,000 inputs as done by \cite{oikarinen2024linear}.

For \textbf{simple explanations}, the predicted activation $s$ is simply the presence of concept on that input.

\begin{equation}
    s(x_i, t) = [c_t]_i
\end{equation}

For a \textbf{linear explanation}, $E = \{(w_1, t_1), ..., (w_l, t_l)\}$ the predicted activation $s$ is calculated following \cite{oikarinen2024linear} as:

\begin{equation}
    s(x_i, E) = \sum_{w_j, t_j \in E} w_j[c_{t_j}]_i
\end{equation}

For \textbf{compositional explanations} \cite{mu2021compositional}, we calculate the predicted activation as follows using probabilistic logic. Different basic logical operators are calculated as:

\begin{equation}
    s(x_i, t_1 \text{ AND } t_2) =  [c_{t_1}]_i \cdot [c_{t_2}]_i
\end{equation}

\begin{equation}
    s(x_i, t_1 \text{ OR } t_2) =  1 - (1-[c_{t_1}]_i) \cdot (1 - [c_{t_2}]_i)
\end{equation}

\begin{equation}
    s(x_i, \text{NOT } t) =  1-[c_{t}]_i
\end{equation}

Predictions for more complex compositions are then calculated by iteratively applying these rules.

\textbf{Clustered Compositional Explanations:} CCE \cite{la2023towards} explanations are of the form $E = \{(l_1, u_1, F_1), ..., (l_r, u_r, F_r)\}$ where $r$ is the number of activation clusters, and $l_j, u_j$ are the lower and upper bound of activations for that cluster and $F_j$ is a compositional explanation for activations of that cluster. To predict neuron activation based on this explanation, we use the following formula:

\begin{equation}
    s(x_i, E) = \sum_{l_j, u_j, F_j \in E} \frac{l_j + u_j}{2}s(x_i, F_j)
\end{equation}

This means if the concepts according to the formula of a cluster are present, we predict the neuron's activation will be in the middle of the clusters activation range.

For all automated evaluations we use SigLIP-SO400M-14-386 model to predict $c_t$ following \cite{oikarinen2024linear}.

\subsection{Theorem 1}
\label{sec:IS_thm}

Suppose we are estimating the expected value of function $h(x)$ when $x\sim\mathcal{P}$. Let $\mathcal{X}$ be the support of $\mathcal{P}$.

\begin{theorem}
    [\cite{montecarlobook}, Sec 3.3.2, Theorem 3.12] For importance sampling with sampling distribution $q$:
    $$
    \mathbb{E}_{x \sim \mathcal{P}}[h(x)] = \int_{\mathcal{X}} h(x) \frac{p(x)}{q(x)}q(x)dx \approx \frac{1}{|S|}\sum_{i=1}^{|S|}\frac{h(x_i)p(x_i)}{q(x_i)}.
    $$
    The choice of $q$ that minimizes the variance satisfies
    $
    q(x) \propto |h(x)|p(x).
    $
    \label{thm:IS}
\end{theorem}

\textbf{Proof:} Reproduced from \cite{montecarlobook}. 

\begin{equation}
    Var\left[\frac{h(x)p(x)}{q(x)}\right] = \mathbb{E}_q\left[\left(\frac{h(x)p(x)}{q(x)}\right)^2\right] -  \left(\mathbb{E}_q\left[\frac{h(x)p(x)}{q(x)}\right]\right)^2
\end{equation}

Since the the second term $\left(\mathbb{E}_q\left[\frac{h(x)p(x)}{q(x)}\right]\right)^2 = \left( \int_{\mathcal{X}} h(x)p(x)dx \right)^2$ does not depend on $q$, in order to minimize variance we only need to minimize the first term.

From Jensen's inequality it follows that:
\begin{equation}
    \mathbb{E}_q\left[\left(\frac{h(x)p(x)}{q(x)}\right)^2\right] \geq \left(\mathbb{E}_q\left[\frac{|h(x)|p(x)}{q(x)}\right]\right)^2
     = \left(\int_{\mathcal{X}} |h(x)|p(x) dx \right)^2
\end{equation}

Giving us a lower bound for the first term. If we set 

\begin{equation}
    q(x) = \frac{|h(x)|p(x)}{\int_{\mathcal{X}} |h(z)|p(z)dz}
\end{equation}

Which is a valid probability distribution, we get:

\begin{equation}
     \mathbb{E}_q\left[\left(\frac{h(x)p(x)}{q(x)}\right)^2\right] = \left(\int_{\mathcal{X}} |h(z)|p(z)dz\right)^2
\end{equation}

This exactly matches the lower bound, proving that minimum variance is attained by setting 

\begin{equation}
    q(x) = \frac{|h(x)|p(x)}{\int_{\mathcal{X}} |h(z)|p(z)dz} \propto |h(x)|p(x)
\end{equation}

\subsection{Compute details}
\label{sec:compute_details}

Our main contribution is focused on efficient crowd-sourced evaluation and as such our method is not computationally costly. The main computational cost associated with our method is calculating the SigLIP image encoder outputs for the entire $\mathcal{D}$, as these are needed for both importance sampling and the Bayes - SigLIP prior. However this is a relatively cheap onetime cost of around 20 minutes on a single NVIDIA RTX 6000 Ada Generation GPU.

The main computational expense associated with this paper involved running the baseline methods. In Tables \ref{tab:runtime_len1} and \ref{tab:runtime_complex} below, we report the approximate runtime to explain all neurons of a layer using different description methods using a single NVIDIA RTX 6000 Ada Generation GPU.

\begin{table*}[h!]
\centering
\scalebox{0.86}{
\begin{tabular}{@{}lcccccccc@{}}
\toprule
\multicolumn{1}{c}{Simple Exp. ($l$=1)} & 
\begin{tabular}[c]{@{}c@{}}ND\\ \cite{bau2020understanding}\end{tabular}
 & \begin{tabular}[c]{@{}c@{}}MILAN\\ \cite{hernandez2022natural}\end{tabular}
  & \begin{tabular}[c]{@{}c@{}}CD\\ \cite{oikarinen2023clip}\end{tabular}
 & \begin{tabular}[c]{@{}c@{}}INVERT\\ l=1~\cite{bykov2023labeling}\end{tabular} & 
 \begin{tabular}[c]{@{}c@{}}DnD\\ \cite{bai2025interpret}\end{tabular}
  & \begin{tabular}[c]{@{}c@{}}MAIA\\ \cite{shaham2024multimodal}\end{tabular}
   & \begin{tabular}[c]{@{}c@{}}LE(label) \\ l=1~\cite{oikarinen2024linear}\end{tabular} & \begin{tabular}[c]{@{}c@{}}LE(SigLIP)\\ l=1~\cite{oikarinen2024linear}\end{tabular} \\ \midrule
\begin{tabular}[c]{@{}l@{}}RN-50 \\ (Layer4)\end{tabular} & $\sim$ 1 hr & $\sim$ 1 hr & $\sim$ 5 mins & $\sim$ 1 hr & $\sim$ 55 hrs & $\sim$ 255 hrs & $\sim$ 1 hr & $\sim$ 1 hr \\ \bottomrule
\\
\end{tabular}}
\caption{Approximate runtime of different $l=1$ baseline methods to explain neurons.}
\label{tab:runtime_len1}
\end{table*}
\begin{table*}[h!]
\centering
\scalebox{0.83}{
\begin{tabular}{@{}lcccccc@{}}
\toprule
\multicolumn{1}{c}{Complex Exp. ($l$>1)} & \begin{tabular}[c]{@{}c@{}}Comp Exp~\cite{mu2021compositional}\\ l=3\end{tabular} & \begin{tabular}[c]{@{}c@{}}INVERT~\cite{bykov2023labeling}\\ l=3\end{tabular} & \begin{tabular}[c]{@{}c@{}}CCE~\cite{la2023towards},\\ l= 5\end{tabular} & \begin{tabular}[c]{@{}c@{}}CCE~\cite{la2023towards},\\ l=15\end{tabular} & \begin{tabular}[c]{@{}c@{}}LE(label)~\cite{oikarinen2024linear}\\ l=4.37/1.97\end{tabular} & \begin{tabular}[c]{@{}c@{}}LE(SigLIP)~\cite{oikarinen2024linear} \\ l=4.66/1.82\end{tabular} \\ \midrule
\begin{tabular}[c]{@{}l@{}}RN-50 \\ (Layer4)\end{tabular} & $\sim 92$ hrs & $\sim 24$ hrs & $\sim 87$ hrs & $\sim 275$ hrs & $\sim 1$ hr & $\sim 1$ hr  \\ \bottomrule \\
\end{tabular}}
\caption{Approximate runtime of different complex explanation baseline methods.}
\label{tab:runtime_complex}
\end{table*}

\subsection{Inter-Annotator Consistency}

To measure the quality of our ratings, we also measure inter-annotator consistency through Fleiss's Kappa. This measures the amount of agreement above chance, with 1 meaning perfect agreement and 0 random chance. 

For MTurk raters in our test Setting 2, we observe a Kappa of 0.395. On the real study we observe Fleiss's Kappa of 0.191 on ViT neuron explanations and 0.274 on RN50 neurons. Overall we can see these agreement rates are relatively low, highlighting the need for error correction trough rating aggregation. While we think some of the noise in responses is underlying poor quality responses, some disagreement is also caused by underlying ambiguity in the task or poor explanations like MAIA sometimes outputting "Undetermined Selectivity".

To compare against higher quality annotations, we had 3 authors rate 150 samples each for the 10 neurons in Setting 2. For authors we observed a Fleiss Kappa of 0.774. This shows that obtaining higher quality raters has potential to improve study signal, but as we show we can still efficiently utilize noisy ratings with our methods.


\end{document}